\documentclass{article}

\usepackage[final]{neurips_2022}

\usepackage{pifont}
\newcommand{\cmark}{\ding{51}}%
\newcommand{\xmark}{\ding{55}}%

\usepackage[utf8]{inputenc} 
\usepackage[T1]{fontenc}    
\usepackage{hyperref}       
\usepackage{url}            
\usepackage{booktabs}       
\usepackage{amsfonts}       
\usepackage{nicefrac}       
\usepackage{microtype}      
\usepackage{xcolor}         

\usepackage{microtype}
\usepackage{graphicx}
\usepackage{booktabs} 

\usepackage{hyperref}
\hypersetup{
    colorlinks,
    citecolor=blue,
    filecolor=black,
    linkcolor=blue,
    urlcolor=blue
}

\usepackage{forloop}
\usepackage[ruled, vlined]{algorithm2e} 
\usepackage{algpseudocode}
\usepackage{wrapfig}
\newcommand{\ah}{$\mathtt{AH}$\xspace}
\newcommand{\ssr}{$\mathtt{SSR}$\xspace}
\newcommand{\ssrrrs}{$\mathtt{SSR}$-$\mathtt{RRS}$\xspace}

\usepackage{mathtools}
\usepackage{csquotes}

\usepackage{amsmath,amsfonts,bm}

\def\eqref#1{equation~\ref{#1}}

\def\1{\bm{1}}
\newcommand{\data}{\mathcal{D}}
\newcommand{\train}{\mathcal{D_{\mathrm{train}}}}
\newcommand{\valid}{\mathcal{D_{\mathrm{valid}}}}

\DeclareMathAlphabet{\mathsfit}{\encodingdefault}{\sfdefault}{m}{sl}
\SetMathAlphabet{\mathsfit}{bold}{\encodingdefault}{\sfdefault}{bx}{n}

\def\gA{{\mathcal{A}}}

\def\gE{{\mathcal{E}}}

\def\gG{{\mathcal{G}}}

\def\gJ{{\mathcal{J}}}

\def\gN{{\mathcal{N}}}

\def\gR{{\mathcal{R}}}
\def\gS{{\mathcal{S}}}
\def\gT{{\mathcal{T}}}

\newcommand{\E}{\mathbb{E}}

\newcommand{\R}{\mathbb{R}}

\DeclareMathOperator*{\argmax}{arg\,max}
\DeclareMathOperator*{\argmin}{arg\,min}

\usepackage{my_definitions}

\usepackage[
  separate-uncertainty = true,
  multi-part-units = repeat
]{siunitx}

\usepackage{amsmath}
\usepackage{amssymb}
\usepackage{mathtools}
\usepackage{amsthm}
\usepackage{bbm}
\usepackage{multirow}

\usepackage[capitalize,noabbrev]{cleveref}

\crefformat{section}{\S#2#1#3} 
\crefformat{subsection}{\S#2#1#3}
\crefformat{subsubsection}{\S#2#1#3}

\theoremstyle{plain}
\theoremstyle{definition}
\theoremstyle{remark}

\usepackage[textsize=tiny]{todonotes}

\title{Data-Efficient Pipeline for Offline Reinforcement Learning with Limited Data}

\author{%
   Allen Nie$^*$ \quad Yannis Flet-Berliac \quad Deon R. Jordan   \\ \textbf{William Steenbergen} \quad \textbf{Emma Brunskill} \vspace{1mm}\\
   Department of Computer Science \\
   Stanford University \\
   *\texttt{anie@stanford.edu}\\
}

\begin{document}

\maketitle



\begin{abstract}
Offline reinforcement learning (RL) can be used to improve future performance by leveraging historical data. There exist many different algorithms for offline RL, and it is well recognized that these algorithms, and their hyperparameter settings, can lead to decision policies with substantially differing performance. This prompts the need for pipelines that allow practitioners to systematically perform algorithm-hyperparameter selection for their setting. Critically, in most real-world settings, this pipeline must only involve the use of historical data. 
Inspired by statistical model selection methods for supervised learning, we introduce a task- and method-agnostic pipeline for automatically training, comparing, selecting, and deploying the best policy when the provided dataset is limited in size. 
In particular, our work highlights the importance of performing multiple data splits to produce more reliable algorithm-hyperparameter selection. While this is a common approach in supervised learning, to our knowledge, this has not been discussed in detail in the offline RL setting. We show it can have substantial impacts when the dataset is small. Compared to alternate approaches, our proposed pipeline outputs higher-performing deployed policies from a broad range of offline policy learning algorithms and across various simulation domains in healthcare, education, and robotics. This work contributes toward the development of a general-purpose meta-algorithm for automatic algorithm-hyperparameter selection for offline RL.

\end{abstract}


\section{Introduction}
\label{sec:intro}

Offline/batch reinforcement learning has the potential to learn better decision policies from existing real-world datasets on sequences of decisions made and their outcomes. In many of these settings, tuning methods online is infeasible and deploying a new policy involves time, effort and potential negative impact. Many of the  existing datasets for applications that may benefit from offline RL may be fairly small in comparison to supervised machine learning. For instance, the MIMIC intensive care unit dataset on sepsis that is  often studied in offline RL has 14k patients~\citep{komorowski2018artificial}, the number of students frequently interacting with an online course will often range from hundreds to tens of thousands~\citep{bassen2020reinforcement}, and the number of demonstrations collected from a human operator manipulating a robotic arm is often on the order of a few hundred per task~\citep{mandlekar2018roboturk}. In these small data regimes, recent studies~\citep{mandlekar2021what,levine2020offline} highlight that with limited data, the selection of hyperparameters using the training set is often challenging. Yet hyperparameter selection also has a  substantial influence on the resulting policy's performance, particularly when the algorithm leverages deep neural networks.


One popular approach to address this is to learn policies from particular algorithm-hyperparameter pairs on a training set and then use offline policy selection, which selects the best policy given a validation set~\citep{pmlr-v37-thomas15,thomas2019preventing,paine2020hyperparameter,kumar2021a}. 
However, when the dataset is limited in size, this approach can be limited: (a) if the validation set happens to have no or very few good/high-reward trajectories, then trained policies cannot be properly evaluated; (b) if the training set has no or very few such trajectories, then no good policy behavior can be learned through any policy learning algorithm; and (c) using one fixed training dataset is prone to overfitting the hyperparameters on this one dataset and different hyperparameters could be picked if the training set changes. One natural solution to this problem is to train on the entire dataset and compare policy performance on the same dataset, which is often referred to as the internal objective approach. In  Appendix~\ref{ap:prelude} we conduct a short experiment using D4RL where this approach fails due to the common issue of Q-value over-estimation~\citep{fujimoto2019off}.

There has been much recent interest in providing more robust methods for offline RL. Many rely on the  workflow just discussed, where methods are trained on one dataset and 
Offline Policy Evaluation (OPE) is used to do policy selection~\citep{su2020adaptive,paine2020hyperparameter,zhang2021towards,kumar2021a,lee2021model,tang2021model,miyaguchi2022theoretical}. Our work highlights the impact of a less studied issue: the challenge caused by data partitioning variance. We first motivate the need to account for train/validation partition randomness by showing the wide distribution of OPE scores the same policy can obtain on different subsets of data or the very different performing policies the same algorithm and hyperparameters can learn depending on different training set partitions. We also prove a single partition can have a notable failure rate in identifying the best algorithm-hyperparameter to learn the best policy. 

%
%




We then introduce a general pipeline for algorithm-hyperparameters (\ah) selection and policy deployment that: (a) uses repeated random sub-sampling (RRS) with replacement of the dataset to perform \ah training, (b) uses OPE on the validation set, (c) computes aggregate statistics over the RRS splits to inform \ah selection, and (d) allows to use the selected \ah to retrain on the entire dataset to obtain the deployment policy. Though such repeated splitting is common in supervised learning, its impact and effect have been little studied in the offline RL framework. Perhaps surprisingly, we show that our simple pipeline leads to substantial performance improvements in a wide range of popular benchmark tasks, including D4RL~\citep{fu2020d4rl} and Robomimic~\citep{mandlekar2021what}.

\begin{table*}[t]
\small
\centering
\begin{tabular}{@{}rccccc@{}}
\toprule
Common Practices                                                                                      & \begin{tabular}[c]{@{}c@{}}Non-Markov \\ Env\end{tabular}  & \begin{tabular}[c]{@{}c@{}}Data \\ Efficient\\(re-train)\end{tabular} & \begin{tabular}[c]{@{}c@{}}Compare \\ Across \\ OPL\end{tabular} & \begin{tabular}[c]{@{}c@{}}Considers \\ Evaluation \\ Variation\end{tabular} & \begin{tabular}[c]{@{}c@{}}Considers \\ Training \\ Variation\end{tabular} \\ \midrule
\multicolumn{6}{l}{\textbf{Policy selection (1 split)}}                                                                                                                                                                                                                                                                                                                                                \\ \midrule
\begin{tabular}[c]{@{}r@{}}Internal Objective / TD-Error\\\citep{thomas2015high,thomas2019preventing}\end{tabular}      & (depends)  &   \xmark                                                        &   \xmark                                                               &   \xmark                                                                           &     \xmark                                                                       \\ \midrule
OPE methods \\ (\cite{komorowski2018artificial};\\\cite{paine2020hyperparameter}                                                                                           &    (depends)                   &   \xmark                                                        &      \cmark                                                            &      \xmark                                                                        &         \xmark                                                                   \\ \midrule
\begin{tabular}[c]{@{}r@{}}OPE + BCa Val. \\\citep{thomas2015high}\end{tabular}                              &  (depends)                                                            &    \xmark                                                       &    \cmark                                                              &     \cmark                                                                         &     \xmark                                                                       \\ \midrule
\begin{tabular}[c]{@{}r@{}} BVFT \\\citep{xie2021batch} \end{tabular}              &  \xmark                                         &   \xmark                                      &  \xmark                                             &  \xmark                                                        &  \xmark                                                       \\ \midrule
\begin{tabular}[c]{@{}r@{}}BVFT + OPE \\\citep{zhang2021towards}\end{tabular}                                                              &      \xmark                     &   \xmark                                                                &    \cmark                                                                          &     \cmark       &  \xmark                                                                 \\ \midrule
\begin{tabular}[c]{@{}r@{}}Q-Function Workflow \\ \citep{kumar2021a}  \end{tabular}                                                                                        &  \xmark                                                             & \cmark                                                          &  \xmark                                                                &    \xmark                                                                          &          \xmark                                                                  \\ \midrule
\multicolumn{6}{l}{\textbf{Ours: $\gA_i$ selection (multi-split)}}                                                                                                                                                                       \\ \midrule
Cross-Validation                                                                            & \cmark                                       & \cmark & \cmark                                      & \cmark                                             & \cmark                                                                                                         \\ \midrule
\begin{tabular}[c]{@{}r@{}}Repeated Random \\ Subsampling\end{tabular}                  & \cmark                                       & \cmark & \cmark                                      & \cmark                                             & \cmark                                                                                                             \\ \bottomrule
\end{tabular}
\caption{A summary of commonly used approaches for choosing a deployment policy from a fixed offline RL dataset. We define \textit{Data Efficient} as: the approach assumes the algorithm can be re-trained on all data points; \textit{(depends)} as: depends on whether the underlying OPL or OPE methods make explicit Markov assumption or not.}
\label{tab:ops}

\end{table*}

\section{Related work}


\looseness=-1\textbf{Offline Policy Learning (OPL).}
In OPL, the goal is to use historical data from a fixed behavior policy $\pi_b$ to learn a reward-maximizing policy in an unknown environment (Markov Decision Process, defined in Section~\ref{sec:problem}).
Most work studying the sampling complexity and efficiency of offline RL~\citep{xie2021batch,yin2021near} do not depend on the structure of a particular problem, but empirical performance may vary with some pathological models that are not necessarily Markovian.~\citet{shi2020does} have precisely developed a model selection procedure for testing the Markovian hypothesis and help explain different performance on different models and MDPs.
To address this problem, it is inherently important to have a fully adaptive characterization in RL because it could save considerable time in designing domain-specific RL solutions~\citep{zanette2019tighter}.
As an answer to a variety of problems, OPL is rich with many different methods ranging from policy gradient~\citep{DBLP:conf/uai/LiuSAB19}, model-based~\citep{yu2020mopo,kidambi2020morel}, to model-free methods~\citep{Siegel2020Keep,fujimoto2019off,guo2020batch,kumar2020conservative} each based on different assumptions on the system dynamics. Practitioners thus dispose of an array of algorithms and corresponding hyperparameters with no clear consensus on a generally applicable evaluation tool for offline policy selection.

\textbf{Offline Policy Evaluation (OPE).}
OPE is concerned with evaluating a target policy's performance using only pre-collected historical data generated by other (behavior) policies~\citep{voloshin2021empirical}. Each of the many OPE estimators has its unique properties, and in this work, we primarily consider two main variants~\citep{voloshin2021empirical}: Weighted Importance Sampling (WIS)~\citep{precup2000eligibility} and Fitted Q-Evaluation (FQE)~\citep{le2019batch}. Both WIS and FQE are sensitive to the partitioning of the evaluation dataset. WIS is undefined on trajectories where the target policy does not overlap with the behavior policy and self-normalizes with respect to other trajectories in the dataset. FQE learns a Q-function using the evaluation dataset. This makes these estimators very different from mean-squared errors or accuracy in the supervised learning setting -- the choice of partitioning will first affect the function approximation in the estimator and then cascade down to the scores they produce.

\textbf{Offline Policy Selection (OPS).}
Typically, OPS is approached via OPE, which estimates the expected return of candidate policies.~\citet{zhang2021towards} address how to improve policy selection in the offline RL setting. The algorithm builds on the Batch Value-Function Tournament (BVFT)~\citep{xie2021batch} approach to estimating the best value function among a set of candidates using piece-wise linear value function approximations and selecting the policy with the smallest projected Bellman error in that space. Previous work on estimator selection for the design of OPE methods include~\citet{su2020adaptive,miyaguchi2022theoretical} while ~\citet{kumar2021a,lee2021model,tang2021model,paine2020hyperparameter} focus on offline hyperparameter tuning.~\citet{kumar2021a} give recommendations on when to stop training a model to avoid overfitting. The approach is exclusively designed for Q-learning methods with direct access to the internal Q-functions. On the contrary, our pipeline does policy training, selection, and deployment on any offline RL method, not reliant on the Markov assumption, and can select the best policy with potentially no access to the internal approximation functions (black box). We give a brief overview of some OPS approaches in Table~\ref{tab:ops}.





\section{Background and Problem Setting}
\label{sec:problem}
We define a stochastic Decision Process $M =$ $\langle \gS, A, T, r, \gamma \rangle$, where $\gS$ is a set of states; $A$ is a set of actions; $T$
is the transition dynamics (which might depend on the full history); $r$
is the reward function; and $\gamma \in (0, 1)$ is the discount factor. Let $\tau = \{s_i, a_i, s_i', r_i\}_{i=0}^L$ be the trajectory sampled from $\pi$ on $M$. The optimal policy $\pi$ is the one that 
maximizes the expected discounted return $V(\pi) = \E_{\tau \sim  \rho_\pi}[G(\tau)]$ where $G(\tau) = \sum_{t=0}^{\infty} \gamma^t r_t$ and $\rho_\pi$ is the distribution of $\tau$ under policy $\pi$.
For simplicity, in this paper we assume policies are Markov $\pi: S \rightarrow A$, but it is straightforward to consider policies that are a function of the full history. In an offline RL problem, we take a dataset: $\data = \{\tau_i\}_{i=1}^n$, which can be collected by one or a group of policies which we refer to as the behavior policy $\pi_b$ on the decision process $M$. The goal in offline/batch RL is to learn a decision policy $\pi$ from a class of policies with the best expected performance $V^{\pi}$ for future use. 
Let $\mathcal{A}_i$ to denote an \ah pair, i.e. an offline policy learning algorithm and its hyperparameters and model architecture. 
%
An offline policy estimator takes in a policy $\pi_e$ and a dataset $\data$, and returns an estimate of its performance: $\widehat V: \Pi \times \data \rightarrow \R$. In this work, we focus on two popular Offline Policy Evaluation (OPE) estimators: Importance Sampling (IS)~\citep{precup2000eligibility} and Fitted Q-Evaluation (FQE)~\citep{le2019batch} estimators. We refer the reader to \cite{voloshin2021empirical} for a more comprehensive discussion.


\begin{figure*}[t]%
\vspace{-2em}
    \centering
    \includegraphics[width=\linewidth]{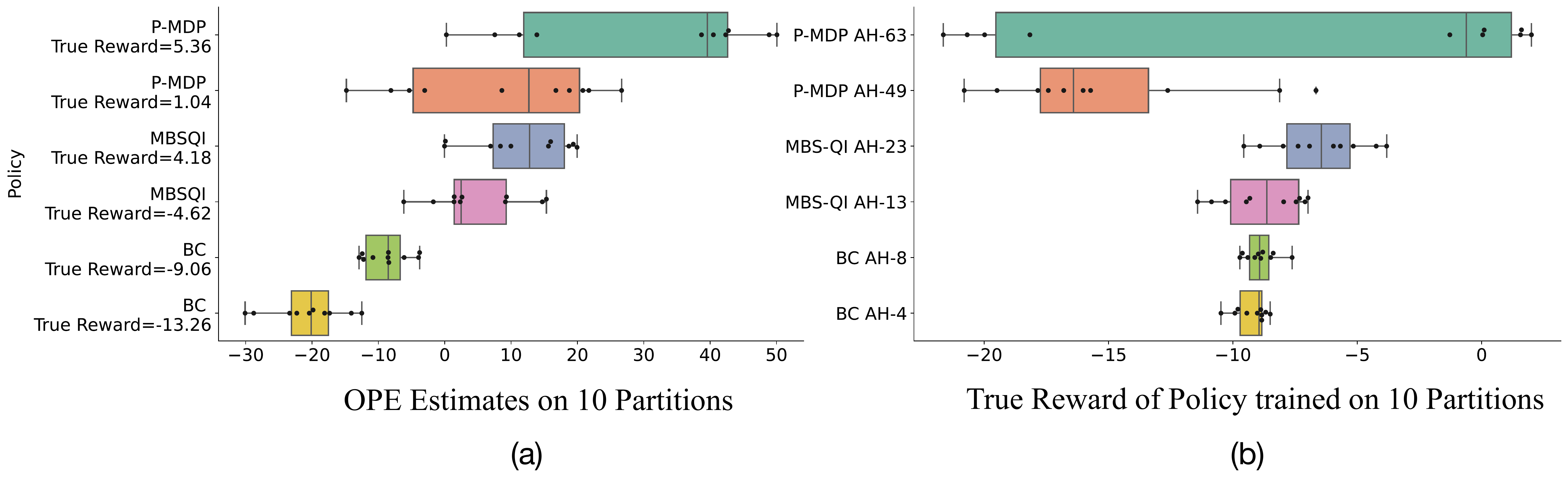}
    \caption{\looseness=-1True performance and evaluation of 6 $\mathcal{A}_i$ pairs on the Sepsis-POMDP (N=1000) domain. (a) shows the OPE estimations and (b) shows the variation in terms of true performance. The variations are due to the different \ah pairs of the policies \textit{but also} to the sensitivity to the training/validation splits.
    }
    \label{fig:new-variation}
\end{figure*}

\section{The Challenge of Offline RL $\mathcal{A}_i$ Selection}
\label{sec:offline}
An interesting use-case of offline RL is when domain experts have access to an existing dataset (with potentially only a few hundred trajectories) about sequences of decisions made and respective outcomes, with the hope of leveraging the dataset to learn a better decision policy for future use. In this setting, the user may want to consider many options regarding the type of RL algorithm (model-based, model-free, or direct policy search), hyperparameter, or deep network architecture to use.

Automated algorithm selection is important because different $\mathcal{A}_i$ (different \ah pairs) may learn very diverse policies, each with significantly different performance $V^{\mathcal{A}_i}$. Naturally, one can expect that various algorithms lead to diverse performance, but using a case-study experiment on a sepsis simulator~\citep{oberst2019counterfactual}, we observe in Figure~\ref{fig:new-variation}(b) that the sensitivity to hyperparameter selection is also substantial (cf. different average values in box plots for each method). For example, MBS-QI~\citep{liu2020provably} learns policies ranging from over -12 to -3 in their performance, depending on the hyperparameters chosen.

Precisely, to address hyperparameter tuning, past work often relies on executing the learned policies in the  simulator/real environment. When this is not feasible, as in many real-world applications, including our sepsis dataset example, where the user may only be able to leverage existing historical data, we have no choice but to rely on off-policy evaluation. Prior work~\citep{thomas2015high,farajtabar2018more,thomas2019preventing,mandlekar2021what} have suggested doing so using a hold-out method, after partitioning the dataset into training and validation sets.

Unfortunately, the partitioning of the dataset itself may result in substantial variability in the training process~\citep{dietterich1998approximate}. We note that this problem is particularly prominent in offline RL where high-reward trajectories are sparse and affect both policy learning and policy evaluation. To explore this hypothesis, we consider the influence of the train/validation partition in the same sepsis domain, and we evaluate the trained policies using the Weighted Importance Sampling (WIS)~\citep{precup2000eligibility} estimator. Figure~\ref{fig:new-variation}(a) shows the policies have drastically different OPE estimations with sensitivity to randomness in the dataset partitioning. We can observe the same phenomena in Figure~\ref{fig:new-variation}(b) with largely different true performances depending on the dataset splitting for most of the policies $\mathcal{A}_i$. This is also illustrated on the left sub-figure of Figure~\ref{fig:heatmap_benefit_split} where in the case where a single train-validation split is used, an $\mathcal{A}_i$ that yields lower-performing policies will often be selected over those that yield higher-performing policies when deployed.

\subsection{Repeated Experiments for Robust Hyperparameter Evaluation in Offline RL}
\label{sec:theory}
We now demonstrate why it is important to conduct repeated random sub-sampling on the dataset in offline RL. Consider a finite set of $J$ offline RL algorithms $\mathcal{A}$. Let the policy produced by algorithm $\mathcal{A}_j$ on training dataset $\mathcal{D}$ be $\pi_j$, its estimated performance on a validation set $\hat{V}^{\pi_j}$, and its true (unknown) value be $V^{\pi_j}.$ Denote the true best resulting policy as $\pi_{j^*} = \argmax_j V^{\pi_j}$ and the corresponding algorithm $\mathcal{A}_{j^*}$. Let the best policy picked based on its validation set performance as $\pi_{\hat{j}^*} = \argmax_j \hat{V}^{\pi_j}$ and the corresponding algorithm $\mathcal{A}_{\hat{j}^*}$. 
\begin{theorem}
\label{thm:benefit_n_splits} 
There exist stochastic decision processes and datasets such that (i) using a single train/validation split procedure that selects an algorithm-hyperparameter with the best performance on the validation dataset will select a suboptimal policy and algorithm with significant finite probability, $P(\pi_{\hat{j}^*} \neq \pi_{j^*}) \geq C$, with corresponding substantial loss in performance $O(V_{max})$, and, in contrast,  (ii) selecting the algorithm-hyperparameter with the best average validation performance across $N_s$ train/validation splits will select the optimal algorithm and policy with probability 1: $\lim_{N_s\to\infty} P(\pi_{\hat{j}^*} = \pi_{j^*}) \to 1.$
\end{theorem}
\textit{Proof Sketch.} Due to space constraints we defer the proof to Appendix~\ref{ap:proof}. Briefly, the proof proceeds by proof by example through constructing a chain-like stochastic decision process and considers a class of algorithms that optimize over differing horizons (see e.g.~\cite{jiang2015dependence,cheng2021heuristic,mazoure2021improving}). The behavior policy is uniformly random meaning that trajectories with high rewards are sparse. This means there is a notable probability that in a single partition of the dataset, the  resulting train and/or validation set may not contain a high reward trajectory, making it impossible to identify that a full horizon algorithm, and resulting policy, is optimal.

In the proof and our experiments, we focus on when the training and validation sets are of equal size. If we use an uneven split, such as $80/20\%$, the failure probability can further increase if only a single partition of the dataset is used. We provide an illustrative example in the Appendix. 
Note that Leave-one-out Cross-Validation (LooCV) will also fail in our setting if we employ, as we do in our algorithm, WIS, because as a biased estimator, WIS will return \textit{the observed return of the behavior policy if averaging over a single trajectory, independent of the target policy to be evaluated}. We explain this further in Appendix~\ref{ap:add-discussion}.





\section{\ssr: Repeated Random Sampling for $\gA_i$ Selection and Deployment}
\label{sec:approach}
\vspace{-3mm}





\looseness=-1 In this paper, we are interested in the following problem: \textit{If offline RL training and evaluation are very sensitive to the partitioning of the dataset, especially in small data regimes, how can we reliably produce a final policy that we are confident is better than others and can be reliably deployed in the real-world?}\\
Instead of considering the sensitivity to data partition as an inherent obstacle for offline policy selection, we view this as statistics to leverage for \emph{$\gA_i$} selection.
We propose a general pipeline: \textit{Split Select Retrain} (\ssr) (of which we provide a pseudo-code in Algorithm~\ref{alg:ssr-rs}, Appendix~\ref{ap:alg}) to reliably optimize for a good deployed policy given only: an offline dataset, an input set of \ah pairs and an off-policy evaluation (OPE) estimator. 
This deployment approach leverages the random variations created by dataset partitioning to select algorithms that perform better \emph{on average} using a robust hyperparameter evaluation approach which we develop below.

First, we split and create different partitions of the input dataset. For each train/validation split, each algorithm-hyperparameter (\ah) is trained on the training set and evaluated using the input OPE method to yield an estimated value on the validation set. These estimated evaluations are then averaged, and the best \ah pair ($\gA^*$) is selected as the one with the highest average score. Now the last step of the \ssr pipeline is to re-use the entire dataset to train one policy $\pi^*$ using $\gA^*$.

\paragraph{Repeated Random Sub-sampling (RRS).}
As Theorem~\ref{thm:benefit_n_splits} suggests, one should ensure a sufficient amount of trajectories in the evaluation partition to lower the failure rate $C$. We propose to create RRS train-validation partitions. 
This approach has many names in the statistical model selection literature, such as Predictive Sample Reuse Method~\citep{geisser1975predictive}, Repeated Learning-Test Method~\citep{burman1989comparative} or Monte-Carlo Cross-Validation~\citep{dubitzky2007fundamentals}. It has also been referred to as Repeated Data Splitting~\citep{chernozhukov2018generic} in the heterogeneous treatment effect literature.\\
We randomly select trajectories in $\data$ and put them into into two parts: $R^{\mathrm{train}}$ and $R^{\mathrm{valid}}$. We repeat this splitting process $K$ times to generate paired datasets: $(R^{\mathrm{train}}_1, R^{\mathrm{valid}}_1), (R^{\mathrm{train}}_2, R^{\mathrm{valid}}_2), ..., (R^{\mathrm{train}}_K, R^{\mathrm{valid}}_K)$. We compute the generalization performance estimate as follows:
\begin{align}
    \gG_{\mathcal{A},\mathrm{RS}_K} = \frac{1}{K} \sum_{k=1}^K \big [\hat V(\mathcal{A}(R^\mathrm{train}_k); R^\mathrm{valid}_k) \big ]
\end{align}

\looseness=-1 A key advantage of overlap partitioning is that it maintains the size of the validation  dataset as $K$ increases. This might be favorable since OPE estimates are highly dependent on the  state-action coverage of the validation dataset -- the more data in the validation  dataset, the better OPE estimators can evaluate a policy's performance. As $K \rightarrow \infty$, RRS approaches the leave-$p$-out cross-validation (CV), where $p$ denotes the number of examples in the validation dataset.
Since there are $\binom{n}{p}$ possible selections of $p$ data points out of $n$ in our dataset, it is infeasible to use exact leave-$p$-out CV when $p > 2$, but a finite $K$ can still offer many advantages. Indeed,~\citet{krzanowski1997assessing} point out that leave-$p$-out estimators will have lower variance compared to leave-one-out estimators, which is what the more commonly used M-fold cross-validation method converges to when $M = n-1$. We discuss more in Appendix~\ref{ap:leave-p-out}.
\section{Experiments}
\label{sec:exp}
\looseness=-1 In this section, we answer the following questions: (a) how does the pipeline \ssrrrs compare to other methods? (b) does the proposed pipeline for $\mathcal{A}_i$ selection and policy deployment allow us to generate the best policy trained on the whole dataset? (c) does re-training on the whole dataset (data efficiency) generate better policies than policies trained on half of the dataset when $\mathcal{A}_i$ is selected by the pipeline? In addition, we conduct two ablation studies to answer to: what number of splits should we use for \ssrrrs, and what is the impact of dataset size on the pipeline results? 

\subsection{Task/Domains}
\looseness=-1 The experimental evaluation involves a variety of real-world and simulated domains, ranging from tabular settings to continuous control robotics environments. We evaluate the performance of \ssr in selecting the best algorithm regardless of task domains and assumptions on task structure. We conduct experiments on eight datasets (Figure~\ref{fig:ssr-tasks}) from five domains (details in Appendix~\ref{ap:domains}), which we give a short description below, and use as many as 540 candidate \ah pairs for the Sepsis POMDP domain.

\begin{figure*}[ht!]%
    \centering
    \includegraphics[width=0.9\linewidth]{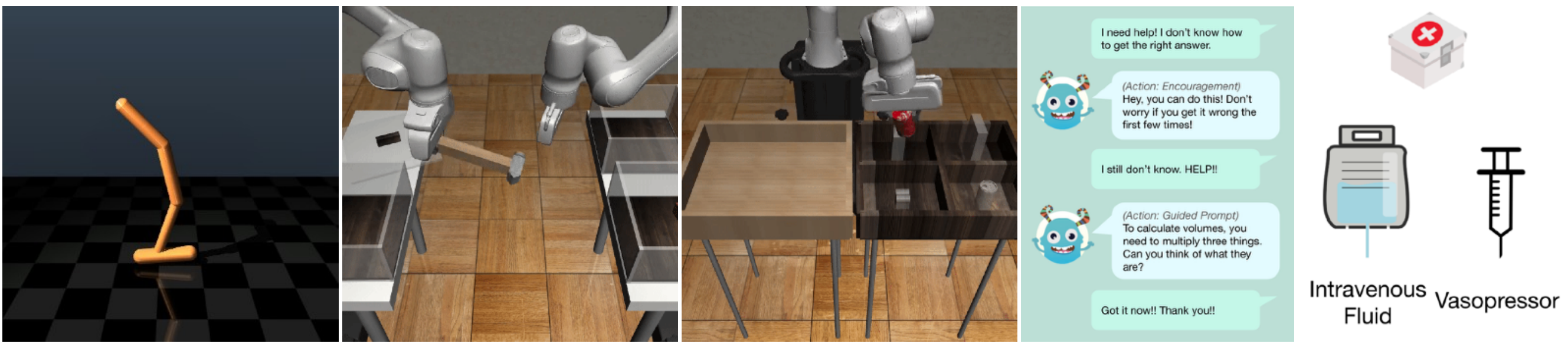}
    \caption{Illustrations from left to right of the D4RL, Robomimic, TutorBot and Sepsis domains.}
    \label{fig:ssr-tasks}
\end{figure*}

\textbf{Sepsis.}
The first domain is based on the simulator and work by \cite{oberst2019counterfactual} and revolves around treating sepsis patients. The goal of the policy for this simulator is to discharge patients from the hospital.
In this domain, we experiment on two tasks: \underline{Sepsis-MDP} and \underline{Sepsis-POMDP}, a POMDP version of Sepsis-MDP.

\textbf{TutorBot.}
The second domain includes a \underline{TutorBot} simulator that is designed to support 3-5th grade elementary school children in understanding the concept of volume and engaging them while doing so. An online study was conducted using a policy-gradient-based RL agent, which interacted with about 200 students.
We took the observations from this online study and built a simulator that reflects student learning progression, combined with some domain knowledge.


\looseness=-1\textbf{Robomimic.}
Robomimic~\citep{mandlekar2021what} is composed of various continuous control robotics environments with suboptimal human data. We use the \underline{Can-Paired} and \underline{Transport} dataset composed of 200 mixed-quality human demonstrations. 
~\citet{mandlekar2021what} attempted to use the RL objective loss on a 20\% split validation set to select the best \ah pair, but reported that the selected \ah did not perform well in the simulator, which makes this task an interesting testbed for our pipeline. 

\looseness=-1\textbf{D4RL.}
D4RL~\citep{fu2020d4rl} is an offline RL standardized benchmark designed and commonly used to evaluate the progress of offline RL algorithms. We use 3 datasets (200k samples each) with different qualities from the Hopper task: \underline{hopper-random} from a randomly initialized policy, \underline{hopper-medium} from a policy trained to approximately 1/3 the performance of a policy trained to completion with SAC ("expert"), and \underline{hopper-medium-expert} from a 50-50 split of medium and expert data.

\begin{figure*}[t]%
\vspace{-2em}
    \centering
    \includegraphics[width=\linewidth]{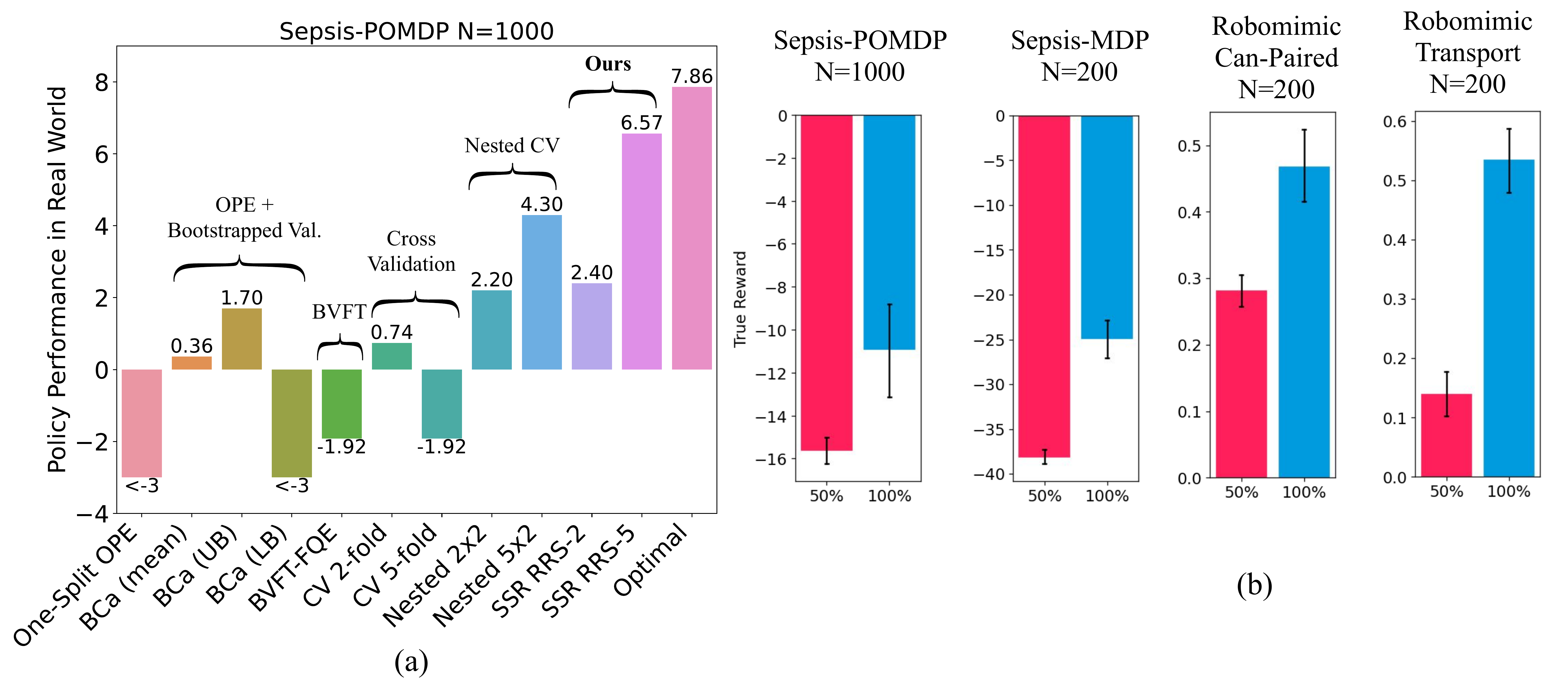}
    \caption{\looseness=-1\textbf{(a)} We first compare our proposed pipeline to various other policy selection approaches in the Sepsis POMDP task. Our approach \ssr-RRS 5-split consistently obtains policies that on average perform close to the optimal policy, significantly outperforming other approaches. \textbf{(b)} We investigate the importance of re-training in the small (N=200) to medium (N=1000) data regime. We show the true reward obtained by all policies from all \ah pairs either trained on 50\% of the data or 100\% of the data. 95\% confidence intervals are depicted as error bars. Most policies achieve higher rewards when trained on more data, even more so when the dataset is small or when tasks are more difficult (Robomimic).}
    \vspace{-1em}
    \label{fig:fig2}
\end{figure*}

\subsection{Baselines}

\textbf{One-Split OPE.} 
\looseness=-1 The simplest method to train and verify an algorithm's performance without access to any simulator is to split the data into a train $\train$ and valid set $\valid$. All policies are trained on the same training set and evaluated on the same valid set. As we explained before, this method has the high potential of overfitting the chosen hyperparameter for one data partition -- it might pick the best policy, but does not guarantee we can use the same hyperparameter to re-train on the full dataset.



\textbf{OPE on Bootstrapped Val.}
Bootstrapping is a popular re-sampling technique that estimates prediction error in supervised learning models~\citep{efron1983estimating,efron1986biased}. The idea of using bootstrapping for OPE estimate is first utilized in HCOPE \citep{thomas2015high}.
Compared to the one-split method and the \ssr pipeline, bootstrapping trains all policies on the same training dataset, and only considers variations in the validation set by creating bootstrapped samples. We refer to the considered Bias-corrected accelerated (BCa) bootstrap method as \textbf{BCa} in the experiments.

\textbf{Cross-Validation (CV).}
One other natural alternative of repeated experiment validation is the popular M-fold Cross-Validation method~\citep{stone1974cross}.
M-fold CV constructs a non-overlapping set of trajectories from the original dataset. For example, a 5-fold CV will train a policy on 80\% of data and evaluate the policy on 20\% of data, as it divides the dataset into 5 non-overlapping partitions.
However, as  we increase the number of splits $M$, which allows us to train/test our algorithms under more data split variations, each non-overlapping set $\mathcal{D}_m$ becomes smaller. When $M=n-1$, M-fold CV becomes leave-one-out CV (LooCV). In this extreme case, many OPE estimators will not work properly, as we have shown in Appendix~\ref{ap:proof}.
We further investigate a variant of M-fold CV called Nested M-fold CV (\textbf{Nested CV}), 
which repeats the M-fold non-overlapping partitioning K times. This procedure is computationally very expensive. Considering the fairness of comparison and computational efficiency, we only evaluate $K \times 2$-fold CV.

\textbf{OPE with BVFT.} Batch Value Function Tournament is the closest competitor to our method, which is a meta-algorithm for hyperparameter-free policy selection~\citep{xie2021batch,zhang2021towards}. For a set of Q-value functions, BVFT makes pairwise comparisons of each (tournament-style) to select the best out of the entire set based on the BVFT-Loss.
Compared to our method, BVFT incurs $\mathcal{O}(J^2)$ comparison given $J$ \ah pairs, practically infeasible for large $J$. The original BVFT can only compare Q-functions, therefore only usable with OPL that directly learns Q-functions. \cite{zhang2021towards} offers an extension to BVFT by using BVFT to compare between FQEs, therefore allowing BVFT to be OPL-agnostic. We adopt the two strategies recommended by the paper. Given $J$ \ah pairs and $B$ FQEs, strategy 1 compares $J \times B$ FQE's Q-functions jointly (\textbf{$\pi$ x FQE}) and strategy 2 compares $B$ FQEs within each \ah and pick the best FQE as the estimate of the \ah's value estimate (\textbf{$\pi$ + FQE}). 
We discuss more in Appendix~\ref{ap:alternative_approaches} (calculations) and \ref{ap:comp_complexity} (time complexity).



\subsection{Training and Evaluation} 
\textbf{Offline Policy Learning.} 
In the following, we outline a variety of Offline RL algorithms used in the evaluation of the \ssr pipeline to demonstrate the generality of our approach and that it can reliably produce the optimal final policy using the selected \ah pair. This marks a departure from workflows designed for specific algorithms, such as~\citet{kumar2021a}. We experiment with popular offline RL methods (see Table~\ref{tab:algs} and we provide algorithmic and hyperparameter details in Table~\ref{tab:domain-details}).
\textbf{Offline Policy Evaluation.}
We use WIS estimators for the tabular and discrete action domains: Sepsis and TutorBot. We use FQE for continuous action domains: Robomimic and D4RL. For each task, with a given dataset, we use the splitting procedure described in Section~\ref{sec:approach} to generate the partitioning. We describe how we compute the results for each figure in Appendix~\ref{ap:figgen}.

\begin{table*}[]
\small
\centering 
\begin{tabular}{@{}rlc@{}}
\toprule
Category                   & \multicolumn{1}{c}{Algorithms}                                         & \multicolumn{1}{c}{\begin{tabular}[c]{@{}c@{}}Internal \\ Q-function\end{tabular}} \\ \midrule
Imitation Learning         & \begin{tabular}[c]{@{}l@{}}BC~\citep{pomerleau1991efficient}\\ BCRNN~\citep{mandlekar2018roboturk}\end{tabular}                     & \xmark                                                                             \\
Conservative Model-Free    & \begin{tabular}[c]{@{}l@{}}BCQ~\citep{fujimoto2019off}, CQL~\citep{kumar2020conservative}, \\ IRIS~\citep{mandlekar2020iris}\end{tabular}              & \cmark                                                                             \\
Policy Gradient & \begin{tabular}[c]{@{}l@{}}POIS~\citep{metelli2018policy}, BC+POIS (ours), BC+mini-POIS (ours)\end{tabular} & \xmark                                                                             \\
Offline Model-Based        & \begin{tabular}[c]{@{}l@{}}MOPO~\citep{yu2020mopo}, P-MDP (ours)\end{tabular}                 & \cmark                                                                             \\ \bottomrule
\end{tabular}
\caption{List of \ah we are comparing in our experiments.}
\label{tab:algs}
\end{table*}

\begin{figure*}[t]%
    \centering
    \includegraphics[width=\linewidth]{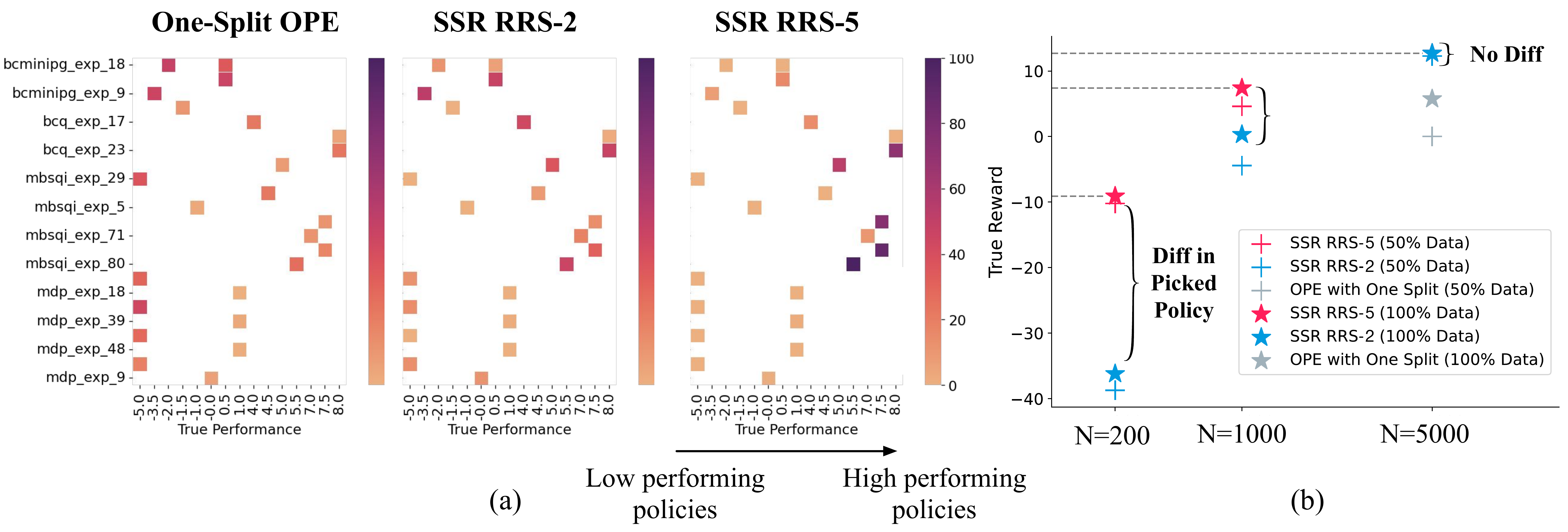}
    \caption{\textbf{(a)} We show the effect of choosing $K$ (the number of train/valid splits) for \ssr-RRS. We ablate $K$ and run the simulation 500 times. The heatmaps shows the frequency at which each policy is chosen by our method and its final performance in the environment. In this experiment, \ssr-RRS chooses over 540 \ah pairs. As we can see, when the number of splits $K$ is larger, \ssr-RRS consistently pick better policies. \textbf{(b)} In the Sepsis-POMDP domain, we show that as the number of trajectories ($N$) in the offline dataset increases, data partitioning becomes less important. Though RRS still outperforms the 1-split policy selection.} 
    \vspace{-1em}
    \label{fig:heatmap_benefit_split}
\end{figure*}

\section{Results}
\label{sec:results}



\begin{table*}[ht!]
\centering
\small
\begin{tabular}{rccccccc}
\toprule
\begin{tabular}[c]{@{}r@{}}Re-trained\\ on full dataset\end{tabular} & \begin{tabular}[c]{@{}c@{}}BVFT-FQE\\ $\pi$ x FQE\end{tabular} & \begin{tabular}[c]{@{}c@{}}BVFT-FQE\\ $\pi$ + FQE\end{tabular} & \begin{tabular}[c]{@{}c@{}} CV-2\end{tabular} & \begin{tabular}[c]{@{}c@{}} CV-5\end{tabular} & \begin{tabular}[c]{@{}c@{}}\ssr\\ RRS-2\end{tabular} & \begin{tabular}[c]{@{}c@{}}\ssr\\ RRS-5\end{tabular} & \multicolumn{1}{c}{\begin{tabular}[c]{@{}c@{}}Optimal\\ Policy\end{tabular}} \\ \midrule
\multicolumn{1}{l}{\textbf{Robomimic:}} &  &  &  &  &  &  &  \\
Can-Paired & \multicolumn{1}{c}{0.65} & \multicolumn{1}{c}{{0.71}} & \multicolumn{1}{c}{{0.72}} & \multicolumn{1}{c}{{0.72}} & \multicolumn{1}{c}{{0.71}} & \multicolumn{1}{c}{\textbf{0.73}} & \multicolumn{1}{c}{0.75} \\
Transport & \multicolumn{1}{c}{0.21} & \multicolumn{1}{c}{0.0} & \multicolumn{1}{c}{0.42} & \multicolumn{1}{c}{0.42} & \multicolumn{1}{c}{0.62} & \multicolumn{1}{c}{\textbf{0.70}} & \multicolumn{1}{c}{0.74} \\ \midrule
\multicolumn{1}{l}{\textbf{D4RL (Hopper):}} &  &  &  &  &  &  &  \\
random & 321.75 & 317.72 & \textbf{325.37} & \textbf{325.37} & 324.92 & \textbf{325.37} & 325.37 \\
medium & 934.71 & 1227.81 & 1227.81 & 1304.53 & 1296.87 & \textbf{1304.54} & 1392.93 \\
medium-expert & 2677.93 & 2677.93  & 2530.04 & 2530.04 & 3481.34 & \textbf{3657.80} & 3657.80 \\ \midrule
\multicolumn{1}{l}{\textbf{Sepsis:}} &  &  &  &  &  &  &  \\
MDP (n=200) & --- & -19.32  & -10.26 & -20.32 & -13.01 & \textbf{-7.85} & -1.94 \\
POMDP (n=1000) & ---  & -1.92 & 0.74 & -1.92 & 2.40 & \textbf{6.75} & 7.86 \\ \midrule
\multicolumn{1}{l}{\textbf{TutorBot:}} &  &  &  &  &  &  &  \\
POMDP (n=200) & ---  & ---  & 1.34 & 1.19 & 1.30 & \textbf{1.38} & 1.43 \\ \bottomrule
\end{tabular}
\caption{Comparison of the performance obtained by a policy deployed using the \ssr pipeline vs. using 1-split policy selection approaches on a wide range of application domains. Cells = average true return. We note that ($\pi$ x FQE) is very computationally expensive when we search through a large \ah space (in Sepsis and TutorBot), therefore we exclude them.}
\label{tab:alldomains}
\vspace{-2mm}
\end{table*}

\textbf{$\mathcal{A}_i$ Selection Comparison.}
In Figure~\ref{fig:fig2}(a), we compare five approaches (One-Split OPE, BCa, BVFT-FQE, CV, and \ssrrrs) on the Sepsis-POMDP domain with 1000 patients. For each approach, we compute a score per \ah pair, and select the best algorithm according to each. For fairness in comparison, all selected $\mathcal{A}_i$ are re-trained on the full dataset and we report the final performance in the real environment. As expected, \textbf{One-Split OPE} performed the worst. Surprisingly, using the lower bound of bootstrapped (\textbf{BCa (LB)}) confidence interval also does not allow to pick good policies, LB being perhaps too conservative. We see that \textbf{CV 2-fold} and \textbf{CV 5-fold} do not perform well either. \textbf{CV 2-fold} does not allow enough repetition and \textbf{CV 5-fold} makes the validation set size too small.
We observe clearly that \textbf{\ssrrrs 5-split} performs the best and selected policies that are on average very close to the optimal policy's performance.
\textbf{BVFT-FQE} relies on FQE, which is a misspecified model on the Sepsis domain and difficult to optimize given the small dataset size; hence it does not select good policies in Sepsis. However, in Robomimic Can-Paired and D4RL Hopper, BVFT-FQE is able to pick good policies, albeit not significantly better or worse than other methods, and still worse than \textbf{\ssrrrs 5-split} in the mixed (more realistic) “medium-expert” dataset. We show more analysis of BVFT compared to our method in the Appendix.
Table~\ref{tab:alldomains} aggregates the results for all the considered domains in our study. Our approach \textbf{\ssrrrs 5-split} is distinctly able to more consistently select policies that, once deployed, perform close to the optimal policy across all tasks.

\textbf{The Benefits of Re-training Policies Selected with \ssr.}
In Figure~\ref{fig:fig2}(b), we plot the true reward of a selected policy $\mathcal{A}_i$ when only trained on 50\% of the dataset (the training set) compared to when trained on 100\% of the dataset. As expected, in the small data regime, every single trajectory matters. Policies trained on the full dataset significantly outperform policies trained only on half of it. This experiment provides strong evidence in favor of \ah selection (done with RRS on the full dataset) over policy selection (done on the training set) in offline RL.

\textbf{The Impact of Number of Repeats for \ssrrrs.}
The proposed pipeline \ssrrrs has a hyperparameter $K$ for the number of repeated data splitting. In Figure~\ref{fig:heatmap_benefit_split}(a), we show the true performance of the policy that is being selected by \textbf{\ssrrrs} with $K=1,2,5$ by running 500 simulations with heatmaps on the frequency of each policy is selected. We observe that when $K=1$ (equivalent to the One-Split OPE method), policies are picked quite uniformly; many of which are performing poorly. When $K=5$, higher-performing policies are selected much more frequently. From Table~\ref{tab:alldomains}, we conclude that $K=5$ generally works well across various domains. Naturally, the number of split $K$ will be chosen in line with the computing budget available; $K=5$ appears to be a reasonable choice.


\textbf{The Impact of Dataset Size.}
\looseness=-1 Finally, we investigate to which extent the proposed pipeline is necessary when the dataset size is sufficiently large. We use the Sepsis-POMDP domain with 200, 1000, and 5000 patients. We show the best policies that are most frequently selected by our approach in Figure~\ref{fig:heatmap_benefit_split}(b). Unsurprisingly, policies trained on larger datasets perform better. In the 200-patient dataset, having \textbf{\ssrrrs 5-split} is crucial in picking the best policy, as most policies perform quite poorly. The gap between different approaches becomes smaller with 1000 patients, and even smaller when there are 5000 patients in the dataset. However, it is worth noting that even in the large dataset regime (N=5000), \textbf{\ssrrrs} still outperforms the One-Split OPE method in selecting the best algorithm.


\textbf{Additional Analysis.} Our method \ssrrrs can also be used to select hyperparameters for a single algorithm, as we demonstrate in Appendix~\ref{ap:robustness}. 
One might also wonder how sensitive is \ssrrrs pipeline to the choice of OPE method used inside the pipeline. OPE methods are known to significantly vary in accuracy for different domains, and unsurprisingly, using a reasonable OPE method for the domain is important (see Appendix~\ref{ap:ope-sensitivity}). Note that the OPE estimators we use in our results are very popular ones, and it is possible to use standard approaches, though additional benefits may come from using even better OPE methods. Finally, related to this question, one might wonder if particular OPE methods might be biased towards certain OPL algorithms which make similar assumptions (such as assuming a Markov structure): interestingly in preliminary experiments, FQE estimators did not seem to give FQI algorithms higher performance estimations (see Appendix~\ref{ap:fqe-fqi-bias}).
\section{Discussion and Conclusion}
\label{sec:discussion}




We presented \ssr, a pipeline for training, comparing, selecting and deploying offline RL policies in a small data regime. The approach performs automated \ah selection with a robust hyperparameter evaluation process using repeated random sub-sampling. \ssr allows to consistently and reliably deploy best-performing policies thanks to jointly avoiding overfitting on a single dataset split and being data efficient in re-using the whole dataset for final training. We prove that a single split has a high failure rate of discovering the optimal \ah because of reward sparsity. We have demonstrated its strong empirical performance across multiple and various challenging domains, including real-world applications where \ah tuning cannot be performed online.

There exist many interesting areas for future work. The proposed offline RL pipeline assumes the user/practitioner has selected a particular OPE method. OPE is an important subarea of its own and different approaches have different bias/variance tradeoffs. Recent work on automated model selection algorithms for OPE~\citep{su2020adaptive,lee2021model} are a promising approach for producing good internal estimators. A second issue is that while our approach aims to produce a high-performing policy, it does not also produce an accurate estimate of this policy since the entire dataset is used at the end for training. An interesting issue is whether cross-splitting~\citep{chernozhukov2016double} or other methods could be used to compute reliable estimators as well as perform policy optimization. 




\section{Acknowledgment}
Research reported in this paper was supported in part by a Hoffman-Yee grant, NSF grant \#2112926 and the DEVCOM Army Research Laboratory under Cooperative Agreement W911NF-17-2-0196 (ARL IoBT CRA). The views and conclusions contained in this document are those of the authors and should not be interpreted as representing the official policies, either expressed or implied, of the Army Research Laboratory or the U.S.Government. The U.S. Government is authorized to reproduce and distribute reprints for Government purposes notwithstanding any copyright notation herein.
We would like to thank Jonathan N. Lee, Henry Zhu, Matthew Jorke, Tong Mu, Scott Fleming, and Eric Zelikman for discussions.




\bibliography{example_paper}
 \bibliographystyle{apalike}

\section*{Checklist}


\begin{enumerate}

\item For all authors...
\begin{enumerate}
  \item Do the main claims made in the abstract and introduction accurately reflect the paper's contributions and scope?
    \answerYes{}
  \item Did you describe the limitations of your work?
    \answerYes{See Section~\ref{sec:results} and Section~\ref{sec:discussion}.}
  \item Did you discuss any potential negative societal impacts of your work?
    \answerNA{}
  \item Have you read the ethics review guidelines and ensured that your paper conforms to them?
    \answerYes{}
\end{enumerate}

\item If you are including theoretical results...
\begin{enumerate}
  \item Did you state the full set of assumptions of all theoretical results?
    \answerYes{See Section~\ref{sec:theory} and Section~\ref{ap:proof}.}
        \item Did you include complete proofs of all theoretical results?
    \answerYes{See Section~\ref{sec:theory} and Section~\ref{ap:proof}.}
\end{enumerate}

\item If you ran experiments...
\begin{enumerate}
  \item Did you include the code, data, and instructions needed to reproduce the main experimental results (either in the supplemental material or as a URL)? \answerYes{See Supplementary Material.}
  \item Did you specify all the training details (e.g., data splits, hyperparameters, how they were chosen)?
    \answerYes{See Sections~\ref{sec:exp},~\ref{ap:domains},~\ref{ap:exp2},~\ref{ap:exp3},~\ref{ap:exp4} and~\ref{ap:exp5}}
    \item Did you report error bars (e.g., with respect to the random seed after running experiments multiple times)?
    \answerYes{}
    \item Did you include the total amount of compute and the type of resources used (e.g., type of GPUs, internal cluster, or cloud provider)?
    \answerYes{}
\end{enumerate}

\item If you are using existing assets (e.g., code, data, models) or curating/releasing new assets...
\begin{enumerate}
  \item If your work uses existing assets, did you cite the creators?
    \answerYes{}
  \item Did you mention the license of the assets?
    \answerNA{Open-source}
  \item Did you include any new assets either in the supplemental material or as a URL?
    \answerYes{}
  \item Did you discuss whether and how consent was obtained from people whose data you're using/curating?
    \answerNA{}
  \item Did you discuss whether the data you are using/curating contains personally identifiable information or offensive content?
    \answerNA{}
\end{enumerate}

\item If you used crowdsourcing or conducted research with human subjects...
\begin{enumerate}
  \item Did you include the full text of instructions given to participants and screenshots, if applicable?
    \answerNA{}
  \item Did you describe any potential participant risks, with links to Institutional Review Board (IRB) approvals, if applicable?
    \answerNA{}
  \item Did you include the estimated hourly wage paid to participants and the total amount spent on participant compensation?
    \answerNA{}
\end{enumerate}

\end{enumerate}

\clearpage
\appendix
\setcounter{figure}{0}
\renewcommand{\thefigure}{A.\arabic{figure}}
\setcounter{table}{0}  
\renewcommand{\thetable}{A.\arabic{table}}
\section{Appendix}

\subsection{Prelude Experiment}
\label{ap:prelude}
In this section, we put ourselves in a situation where model selection would be performed by comparing different \ah pairs on their internal objective or value function estimates on a given dataset, as described near the beginning of Section~\ref{sec:intro}. We use three datasets of different qualities (random, medium, and medium-expert) of the popular Hopper task from the D4RL benchmark (see Appendix~\ref{ap:d4rl} for a detailed description) to train a total of 36 policies with different \ah pairs and then calculate the resulting TD-Errors and Q-values on the whole dataset at the end of training.

To evaluate the performance one would obtain by employing such an approach to select the best policy, we report in Table~\ref{tab:prelude-exp} the performance (true return in the environment) of the selected policies and compare them with the performance of the optimal policy for each of the datasets. The policies are selected either by finding the one which corresponds to the lowest TD-Error, or the one which corresponds to the highest Q-value. We also include the Kendall rank correlation coefficient~\citep{gilpin1993table} for each of the ranking methods (ranking with respect to TD-Error or Q-value) compared with the “true ranking” of policies ranked with respect to the performance in the environment:
$$\tau=\frac{(\text { number of concordant pairs })-(\text { number of discordant pairs })}{\left(\begin{array}{l}
n \\
2
\end{array}\right)}$$ where n is the number of policies, and where “concordant pairs” are pairs from the two compared rankings for which the sort order agrees. A coefficient of 1 means the agreement between the two rankings is perfect.

\begin{table*}[ht!]
\centering
\begin{tabular}{@{}rccccc@{}}
\toprule
                       & \multicolumn{2}{c|}{TD-Error}                                                                                    & \multicolumn{2}{c|}{Q-value}                                                                                     &                                                                                 \\ \midrule
                       & \textbf{\begin{tabular}[c]{@{}c@{}}Policy Selected\\ (True Return)\end{tabular}} & \multicolumn{1}{c|}{Kendall} & \textbf{\begin{tabular}[c]{@{}c@{}}Policy Selected\\ (True Return)\end{tabular}} & \multicolumn{1}{c|}{Kendall} & \textbf{\begin{tabular}[c]{@{}c@{}}Optimal Policy\\ (True Return)\end{tabular}} \\ \midrule
\textbf{random}        & 334.24                                 & \multicolumn{1}{c|}{-0.09}   & 333.65                                                                           & \multicolumn{1}{c|}{-0.15}   & 345.39                       \\
\textbf{medium}        & 1475.82                              & \multicolumn{1}{c|}{0.42}    & 2381.37                                                                          & \multicolumn{1}{c|}{0.21}    & 2469.81                                 \\
\textbf{medium-expert} & 327.97                             & \multicolumn{1}{c|}{-0.18}                     & 327.97                                                                           & \multicolumn{1}{c|}{-0.09}                      & 3657.80                         \\ \bottomrule
\end{tabular}
\caption{Average Return (True Return obtained in the simulator) of the policy selected with respect to min(TD-Error) or max(Q-value) on the training dataset with a comparison to the True Return obtained by the Optimal Policy. Kendall rank correlation coefficient when ranking with respect to the same metrics. Policies are \textbf{trained and validated on the same dataset}. Task: Hopper.}
\label{tab:prelude-exp}
\end{table*}

Unsurprisingly, Table~\ref{tab:prelude-exp} shows that one cannot rely on this straightforward pipeline to select a  best-performing \ah pair. Actually, for most of the datasets (the medium-expert dataset should resemble the most to what a dataset would look like in a real-world situation as it is composed of both high-quality and medium-quality data), following such an approach would produce and deploy a very bad performing policy.

\subsection{Connection between Leave-p-Out CV and RRS}
\label{ap:leave-p-out}

Our RSS is a finite approximation of Leave-p-out (Lp0) cross-validation\footnote{\url{https://scikit-learn.org/stable/modules/generated/sklearn.model_selection.LeavePOut.html}}. LpO is known in supervised learning, but rarely used due to the computational burden. The correctness of LpO is proved ite{celisse2014optimal} in a supervised learning setting with projection estimators. Unlike K-fold cross-validation, Leave-p-out CV selects $p$ data points for evaluation and the rest for training. In our proposed RSS method, we set $p = n / 2$, and instead of exhaustively enumerating all possible selections of $p$ data points out of n data points, we only repeat this process $K$ times. Asymptotically as the amount of data goes to infinity, this approach should be correct, but also a single train/test split will also be correct in such a setting. The key challenges arise in the finite data setting, where the choice of dataset partitioning is key. 

\subsection{Proof of Theorem~\ref{thm:benefit_n_splits}}
\label{ap:proof}
Consider a finite set of $J$ offline RL algorithms $\mathcal{A}$. Let the policy produced by algorithm $\mathcal{A}_j$ on training dataset $\mathcal{D}$ be $\pi_j$, its estimated performance on a validation set $\hat{V}^{\pi_j}$,  and its true (unknown) value be $V^{\pi_j}.$ Denote the true best resulting policy as $\pi_{j^*} = \argmax_j V^{\pi_j}$ and the corresponding algorithm $\mathcal{A}_{j^*}$. Let the best policy picked based on its validation set performance as $\pi_{\hat{j}^*} = \argmax_j \hat{V}^{\pi_j}$ and the corresponding algorithm $\mathcal{A}_{\hat{j}^*}$.

\textbf{Theorem 1. }\textit{Then there exist stochastic decision processes and datasets such that (i) using a single train/validation split procedure will select a suboptimal policy and algorithm with significant finite probability, $P(\pi_{\hat{j}^*} \neq \pi_{j^*}) \geq C$, with corresponding substantial loss in performance $O(V_{max})$, and, in contrast,  (ii) averaging across $N_s$ train/validation splits will select the optimal policy with probability 1: $\lim_{N_s\to\infty} P(\pi_{\hat{j}^*} = \pi_{j^*}) \to 1.$}
\begin{proof}
We proceed by constructing a stochastic decision process. A common domain to illustrate the importance of strategic exploration is a chain MDP. Here consider an episodic, finite horizon, finite chain, deterministic decision process with 6 states, $s_1,\ldots,s_H$, ($H=6$) with two actions. $a_1$ moves the state one down except for at the starting state, and $a_2$ increments the state one up except for the final state: more formally,  $p(s_{i-1}|s_i,a_1)=1$ except for $p(s_1|s_1,a_1)=1$; $p(s_{i+1}|s_i,a_2)$ except for $p(s_H|s_H,a_2)=1$. The reward is 0 in all states except $R(s_1)=1/6$ and $R(s_H)=201$. All episodes are length $H=6$  and start in state $s_1$. The optimal policy always takes $a_2$ and achieves $V_{max}=R(s_H).$ Any other policy achieves at most $H*1/6=1$ reward.  

The behavior policy is uniform random over the two actions, $\pi_b(a_1|s)=0.5=\pi_b(a_2).$ Let the available offline dataset $D$ consist of 200 episodes gathered using $\pi_b.$ Given the behavior policy, each of the 64$=2^H$ unique trajectories has an equal probability of being observed, and only one of these $\tau_h=(s_1,0,a_2,s_2,0,a_2,,s_3,0,a_2,s_4,0,a_2,s_5,a_2,s_H,R(s_H))$ achieves the highest return. On average out of 200 episodes\footnote{Our calculations can easily be extended to cases where there are different numbers of observed $\tau_h$, but for simplicity we assume a dataset where the average expected number of $\tau_h$ are observed.}, $n_{\tau_h}$=3(=$round(|\mathcal{D}|/(2^H))$) episodes will match $\tau_h$. All other episodes will have a return of 1 or less.

Let there be a set of $H$ offline RL algorithms $\mathcal{A}_h$, each which optimizes the reward over a different horizon $h=1:H$, by constructing a maximum-likelihood estimate (MLE) MDP model $\mathcal{M}$ given a training dataset $D_{tr}$, and then computing a policy $\pi_h$ that optimizes the $h$-step value given the learned MDP  $\mathcal{M}$ model\footnote{During planning with the learned MDP model, we restrict taking the maximum value over actions for a given state $s$ to only actions that have been taken at least once in that state in the dataset, e.g. $\max_{a \,s.t.\, n(s,a) \geq 1}$, where $n(s,a)$ is the counts of the number of times action $a$ was taken in state $s$ in the dataset. Note that in a finite dataset, some states and/or actions may not be observed, and this common choice simply ensures that the algorithm does not overestimate the value of untried actions.}
For example, algorithm $\mathcal{A}_2$ will take the MLE MDP model and construct a policy to optimize the sum over rewards for the next two time steps $\pi_2(s) = \arg\max_a r(s) + \sum_{s'} p(s'|s,a) r(s').$ We think this is a reasonable set of algorithms to consider as an illustrative example: the horizon length can directly influence the amount of data needed to compute an optimal policy, and recent work has explored using shorter horizons~\citep{cheng2021heuristic,mazoure2021improving,liao2020personalized}, so choosing the right horizon can be viewed as a bias/variance tradeoff, suitable for automatic model selection. 

Observe that even if given access to the true (unknown) MDP parameters, algorithms $\mathcal{A}_1,\ldots,\mathcal{A}_{H-1}$ will compute a policy that is suboptimal: due to the shortened horizon length, to optimize the expected total reward, the resulting policy computed for $s_1$ will be $\pi_h(s_1)=\pi_1(s_1)=a_1$ for these algorithms $\mathcal{A}_h$, $h=1:{H-1}$. As the MDP is deterministic, this will also be true for any input dataset. 


We now consider the impacts of partitioning the input dataset into a training dataset $D_{tr}$ taken as input by each algorithm $\mathcal{A}_h$ to compute a policy $\pi_{\hat{h}}$, and an evaluation/test dataset $D_{te}$: $D= D_{tr} \cup D_{te}$. For algorithm  $\mathcal{A}_H$ to learn the optimal policy $\pi^*$ which achieves $V_{max}$, it must learn over a dataset $D_{tr}$ that includes one or more examples of the highest return trajectory $\tau_h$. Note that a single episode of $\tau_h$ in the training set is sufficient to learn the optimal policy\footnote{A single example of $\tau_h$ will induce a MLE $\hat{\mathcal{M}}$ with the correct reward model for all states, and the dynamics model for action $a_2$. From the procedure used to compute an optimal policy $\hat{\mathcal{M}}$, this will result in an optimal policy.}.

Assume that the offline evaluation of the policies learned by the algorithm on $D_{tr}$ is performed using importance sampling on $D_{te}$: note, our results will still apply, with minor modifications, if off policy evaluation is performed on $D_{te}$ using fitted Q evaluation~\citep{le2019batch} or using a certainty-equivalent MDP constructed from $D_{te}$. 

Then the off policy evaluation of the policy learned by the full horizon algorithm $\mathcal{A}_H$, $\hat{V}^{\pi_H}(s_1)$, will only be greater than 1 if there also exists at least one episode of the highest return trajectory $\tau_h$. 

Assume the training dataset and validation dataset are constructed by randomly sampling 50\% of the episodes to be in each. By assumption, there are $n_{\tau_h}$ samples of $\tau_h$, which have an equal chance of being in either the training or validation set. There are $n_{\tau_h}+1$ ways to partition the $n_{\tau_h}$ exchangable episodes of $\tau_h$ into the training and validation sets, here $([3,0],[2,1],[1,2],[0,3])$. Note the training and validation set are identical in size ($|\mathcal{D}|/2$ trajectories each), and we only care about whether a trajectory $\tau$ is identical to $\tau_h$ or not. The probability that each of these partitions occurs is : $P([3,0]) = P([0,3]) = \frac{100}{200}*\frac{99}{199}*\frac{98}{198}\approx 0.123$. 

From the above analysis, $\mathcal{A}_H$ can only learn an optimal policy, and its estimated value $\hat{V}^{\pi_5}> 1$ on $D_{te}$ if there is at least one $\tau_h$ in both the training and validation set datasets, which occurs in partitions $([2,1],[1,2])$. This occurs with probability $0.754$. Otherwise, either (a) $\mathcal{A}_H$ will not learn an optimal policy, and instead will learn $\pi_H(s_1) = \pi_1(s_1)=a_1$, or (b) $\mathcal{A}_H$ will learn an optimal policy $\pi_H(s_1)=a_2$ but as the validation dataset does not contain  $\tau_h$, $\hat{V}^{\pi_H} = 1/H < \hat{V}^{\pi_1}$. In both cases, the selected policy given its performance on the validation set will be $\pi_1(s_1)=a_1$. The resulting loss in performance is $V_{max} - V^{\pi_1} = V_{\max} - 1 = O(V_{\max}).$ This failure occurs with substantial probability $24.6\%$. This proves part (i) of the theorem. 

To prove part (ii) of the proof, we consider cases where at least one $\tau_h$ is in both $D_{tr}$ and $D_{te}$.  Note 
$\hat{V}^{\pi_1} \leq \frac{R(s_1)}{1/2^H} = \frac{1}{1/2^H}$.  Define $E_{ss}$ as a "successful split": the event that 1 or more of $\tau_h$ (high returns) episodes are in $D_{te}$, but not all $n_{\tau_h}$. On event $E_{ss}$,  the optimal policy (which will be computed by $\mathcal{A}_H$ on the training set), will have an estimated value on $D_{te}$, using importance sampling:
\begin{equation}
\hat{V}^{\pi^*}_{ss} \geq \frac{1}{|D_{te}|} \frac{R(s_H)}{1/2^H}  = \frac{1}{1/2^H}*\frac{201}{100} > 2 \hat{V}^{\pi_1} \label{eqn:vpi1}
\end{equation}
since there are at least $1$ $\tau_h$ trajectories, each with propensity weight $\frac{1}{1/2^H}$ and reward $R(s_H)$. 
Therefore on Event $E_{ss}$ the optimal policy can be learned and estimated as having high reward. The probability of event $E_{ss}$ is greater than 0.5: $P(E_{ss}) = 0.754.$ 

In the repeated train-validation split setting, the algorithm selected is the one that has the best performance on the validation set, on average across all $N_s$ splits. Let $E_{h}$ be the event that at least half the train-validation dataset splits are successful (Event $E_{ss}$ holds for that split). In this case then the average performance of $\mathcal{A}_5$ will be at least
\begin{eqnarray*}
\hat{V}_{\mathcal{A}_5} &\geq& \frac{1}{N_s} \left( \frac{N_s}{2} \hat{V}^{\pi^*}_{ss} + 0 \right) \\
&\geq& \frac{1}{N_s} \left( \frac{N_s}{2} 2 \hat{V}^{\pi_1} + 0 \right) \\
&=& \hat{V}^{\pi_1},
\end{eqnarray*}
where the first line uses a lower bound of 0 when the event $E_{ss}$ fails to hold, and substitutes in Equation~\ref{eqn:vpi1}. Therefore as long as event $E_h$ holds, the optimal policy $\pi^*$ (which will be computed by algorithm $\mathcal{A}_H$ will be selected. Since $P(E_{ss}) > 0.5$, the probability\footnote{**calculate for finite S.} as the number of splits goes to infinity that $E_{ss}$ holds on least half of those splits goes to 1: $\lim_{N_s \to \infty} P (E_h) \to 1$.



%





\vfill
\end{proof}



\subsection{\ssr pseudo-code}
\label{ap:alg}

\noindent\makebox[\textwidth][c]{
\begin{minipage}{0.55\textwidth}
\centering
\begin{algorithm}[H]
\SetAlgoLined
\SetKwFunction{subsample}{Subsample}
\SetKwFunction{append}{append}
\KwIn{offline RL data $\data$; set of AH pairs [$\gA_1, \gA_2, ..., \gA_z$], OPE estimator  $\widehat V$, split number $K\in \mathbb{N}$.}
\KwOut{policy $\hat \pi^*$ for deployment}
\BlankLine
$\gR$ = $\emptyset$\\ 
\For{$i \leftarrow 1...K$}{
    $R^\mathrm{train}_i, R^\mathrm{valid}_i = $ \subsample($\data, 0.5$) \\
    $\gR = \gR \cup (R^\mathrm{train}_i, R^\mathrm{valid}_i)$
}
$\gG$ = []  \\ 
\For{$i \leftarrow 1...z$}{
    $\gS$ = [] \\
    \For{$j \leftarrow 1...K$}{
        $\pi_i = \mathcal{A}(R^\mathrm{train}_j)$ \\
        $\gS_{ij} = \widehat V(\pi_i; R^\mathrm{valid}_j)$ \\
     }
     $\gG_i = \frac{1}{K} \sum_{j=1}^K \gS_{ij}$
 }
  $\gA^* = \gA_{o+}$ where $o = \argmax(\gG)$\\
 $\pi^* = \gA^*(\data)$  \\ 
 \Return{$\pi^*$}
 \caption{\ssrrrs: $\gA_i$ Selection with Repeated Random Sub-sampling}
 \label{alg:ssr-rs}
\end{algorithm}

\end{minipage}
}

\subsection{Code}

We include the implementation and experiment code here: 
\url{https://github.com/StanfordAI4HI/Split-select-retrain}

\subsection{Experiment Detail Summary}
\label{ap:alternative_approaches}

We choose different sets of algorithms to evaluate our pipeline in every domain to demonstrate the generality of our approach and because some algorithms have limitations inherent to certain types of domains to which they can be applied. We list them in Table~\ref{tab:domain-details}.


\begin{table*}[h]
\centering
\begin{tabular}{@{}rccccl@{}}
\toprule
\begin{tabular}[c]{@{}c@{}}Experiment \\ Domain \end{tabular}              & \begin{tabular}[c]{@{}c@{}}Number of \\ Trajectories (N)\end{tabular} & \begin{tabular}[c]{@{}c@{}}Average \\ Trajectory \\ Length \end{tabular} & \begin{tabular}[c]{@{}c@{}}Number of  \\ Transitions \\ in Total\end{tabular} & \begin{tabular}[c]{@{}c@{}}AH Pairs \\ Evaluated\end{tabular} & \begin{tabular}[c]{@{}c@{}} Algorithms in \\ Experiment \end{tabular}                                                                              \\ \midrule
Sepsis-POMDP                                                     & 200                                                                  & 14                        & 2792                                                                      & 540                                                           & \begin{tabular}[c]{@{}l@{}}BC, POIS, BC+POIS, \\ BC+mini-POIS, BCQ, \\ MBSQI, pMDP, MOPO\end{tabular} \\ \midrule
Sepsis-POMDP                                                     & 1000                                                                 & 14                        & 13708                                                                     & 540                                                           & \begin{tabular}[c]{@{}l@{}}BC, POIS, BC+POIS,\\ BC+mini-POIS,  BCQ, \\ MBSQI, pMDP, MOPO\end{tabular} \\ \midrule
Sepsis-POMDP                                                     & 5000                                                                 & 14                        & 68576                                                                     & 148                                                           & \begin{tabular}[c]{@{}l@{}}BC, POIS, BCQ, \\ MBS-QI, pMDP, MOPO \end{tabular}                         \\ \midrule
Sepsis-MDP                                                       & 200                                                                  & 14                        & 2792                                                                      & 383                                                           & \begin{tabular}[c]{@{}l@{}}BC, BCQ, MBSQI, pMDP, \\ POIS, BC + POIS \\ BC+mini-POIS\end{tabular}      \\ \midrule
TutorBot                                                         & 200                                                                  & 5                         & 987                                                                       & 81                                                            & \begin{tabular}[c]{@{}l@{}}BC, POIS, BC+POIS, \\ BC+mini-POIS\end{tabular}                            \\ \midrule
\begin{tabular}[c]{@{}r@{}}Robomimic \\ Can-Paired \end{tabular} & 200                                                                  & 235                       & 47,000                                                                       & 35                                                            & \begin{tabular}[c]{@{}l@{}}BC, BCRNN, CQL, \\ IRIS, BCQ\end{tabular}\\ \midrule
\begin{tabular}[c]{@{}r@{}}Robomimic \\ Transport \end{tabular}  & 200                                                                  & 470                       & 94,000                                                                       & 10                                                            & \begin{tabular}[c]{@{}l@{}}BC, BCRNN, CQL, \\ IRIS, BCQ\end{tabular}                                  \\ \midrule
\begin{tabular}[c]{@{}r@{}}D4RL\\Hopper\end{tabular} & 500            & 1000      & 500,000         & 4 x 4   & \begin{tabular}[c]{@{}l@{}}BCQ\end{tabular}      \\ \midrule 
\begin{tabular}[c]{@{}r@{}}D4RL\\HalfCheetah\end{tabular} & 500            & 1000      & 500,000         & 4 x 4   & \begin{tabular}[c]{@{}l@{}}BCQ\end{tabular}      \\ 
\bottomrule
\end{tabular}
\caption{List of algorithms being used in which domain. 4 x 4 means we evaluate 4 AH pairs for the policy learning and 4 AH pairs for the policy evaluation estimators (FQE).}
\label{tab:domain-details}
\end{table*}

Running a large number of algorithm-hyperparameter pairs many times is very computationally expensive. In order to save time and resources, we leverage the fact that multiple approaches can share resources. We describe how we compute the numbers for each approach as follows:

For each offline RL dataset in Sepsis, TutorBot, Robomimic, and D4RL, we produce the following partitions (we refer to this as the ``partition generation procedure''):
\begin{enumerate}
    \item 2-fold CV split (2 partitions consisted of $(S_i)$)
    \item 5-fold CV split (5 partitions consisted of $(S_i)$)
    \item 5 RRS split (5 partitions consisted of $(R^{\mathrm{train}}_i, R^{\mathrm{valid}}_i)$)
\end{enumerate}

Here, we briefly describe how to use these data partitions to select algorithms with alternative approaches.

\paragraph{One-Split OPE.} The One-Split OPE method can be conducted to train and evaluate an algorithm on any of the RRS splits being produced, but only look at one split, without considering other splits. We let for a particular $i$, we let $\train = R^{\mathrm{train}_i}$ and $\valid = R^{\mathrm{valid}}_i$.


\paragraph{BCa Bootstrap.}
Similar to the One-Split OPE method, we can use RRS split for bootstrap. For a particular $i$, we let $\train = R^{\mathrm{train}_i}$ and $\valid = R^{\mathrm{valid}}_i$. Bootstrapping will re-sample with replacement on trajectories in $\valid$ to create (largely) overlapping subsets $B_1, B_2, ..., B_N$, with $|B_i|=n$. We then evaluate $\pi_e$ on each subset using $\widehat V$. The final score is computed through a bias correction process with an added acceleration factor (BCa).

\paragraph{Nested K $\times$ 2-fold Cross-Validation.}
We can also use the RRS split partitions to produce $K \times 2$ Nested CV by taking one RRS split $(R^{\mathrm{train}}_i, R^{\mathrm{valid}}_i)$ by doing the following procedure: 
\begin{align}
    &s_i = \frac{\widehat V(\gA(R^{\mathrm{train}}_i); R^{\mathrm{valid}}_i) + \widehat V(\gA(R^{\mathrm{valid}}_i);R^{\mathrm{train}}_i)}{2} \label{eq:2-fold} \\
    &\gG_{\gA, \mathrm{NCV}_K} = \frac{1}{K} \sum_{i=1}^K s_i
\end{align}
Intuitively, for $K \times 2$ Nested CV, we just need to swap the train and valid set produced by repeated sub-sampling and average to produce the algorithm performance score for a particular split $i$. Then we average the scores to get a final score for the algorithm.

\paragraph{2-fold Cross-Validation.} Similar to the $K \times 2$ Nested CV, we can choose the $i$-th partition generated by the 10 RRS split procedure, and compute the score according to Equation~\ref{eq:2-fold}. We do this for the Sepsis and TutorBot domains, but we do not do this for the Robomimic domain.

\paragraph{Batch Value Function Tournament (BVFT)} \cite{xie2021batch,zhang2021towards} proposed to use pairwise Q-function comparisons to select the optimal Q-function from a set of Q-functions. Given $Q_i, Q_j$, let $\gG_{ij}$ be the piecewise constant function class induced by binning $(s, a)$ and $(s', a')$ if $Q_i(s, a) = Q_j(s', a')$. Given an offline dataset $D$, we can compute the BVFT loss as follow:
\begin{align}
    &\hat \gT_{\gG_{ij}} Q \coloneqq \argmin_{g \in \gG_{ij}} \frac{1}{|D|} \sum [(g(s, a) - r - \gamma \max_{a'} Q(s', a'))^2] \\
    &\gE_{\epsilon_k}(Q_i, Q_j) = \|Q_i - \hat \gT_{\gG_{ij}} Q_j \|_{2, D} \\
    &\gE_{\epsilon_k}(Q_i) = \max_j \gE_{\epsilon_k}(Q_i, Q_j)
\end{align}

\cite{zhang2021towards} proposed a method to automatically search through different discretization resolutions ($\epsilon_k$). In our experiment, we search through $[0.1, 0.2, 0.5, 0.7, 1.0, 3.0, 10.0]$. We use the BVFT code provided by~\cite{xie2021batch}.
Because BVFT can only compare Q-functions, \cite{zhang2021towards} offered two strategies to perform policy selection for any model/algorithm. Here we briefly describe two strategies:

\begin{itemize}
    \item Strategy 1 ($\pi$ x FQE): if we have 4 policies, and each policy is evaluated by 4 FQEs, then this strategy will compare 16 Q-functions (4 $\pi$ x 4 FQE). 
    \item Strategy 2 ($\pi$ + FQE): if we have 4 policies, and each policy is evaluated by 4 FQEs, then this strategy will first run BVFT to compare 4 Q-functions (1 $\pi$ x 4 FQE), select the best Q-function for each $\pi$ (4 $\pi$ x 1 FQE), then we select the best policy by the average Q-value computed by each FQE.
\end{itemize}

We generally find strategy 2 more computationally efficient (because it makes a smaller number of comparisons). BVFT generally has $O(J^2)$ time complexity where $J$ is the number of Q-functions that need to be compared -- it's easy to see that $16^2 = 256$ is much larger than $4^2 = 16$. 

Our repeated experiment protocol (RRS) is reliant on choosing a good FQE. In order to compare fairly, for $\pi$ x FQE strategy, we only use the optimal FQE (the ones used in RRS and CV and one-split). We can see that in this condition, BVFT can do pretty well (even outperforming RRS in the D4RL-Hopper medium setting). For $\pi$ + FQE, because it focuses on the selection of FQE, we try 4 different FQE hyperparameters. We discuss this more in D4RL Experiment Details (in Section~\ref{ap:exp5}).

\subsection{Computational Complexity}
\label{ap:comp_complexity}

Most of the approaches we discussed in Section~\ref{ap:alternative_approaches} leverage multiple repetitions (resampling) to account for data allocation randomness. We provide a time complexity table below and define the following terms:
\begin{itemize}
    \item H = number of AH pairs to evaluate
\item N = total data samples. We assume the training time for each trajectory is $N_1$ and evaluation time for each trajectory is $N_2$, where $N = N_1 + N_2$
\item M = number of folds in multi-fold cross-validation
\item B = number of bootstraps (this number is 100 in our experiment)
\item P = number of resolutions for BVFT’s grid (proposed in \cite{zhang2021towards})
\item F = number of FQE hyperparameters (proposed in \cite{zhang2021towards})
\end{itemize}

\begin{table}[h]
\centering
\begin{tabular}{@{}lll@{}}
\toprule
& Training Complexity                                                                & Evaluation Complexity                                                           \\ \midrule
One-Split               & H $\times$ $N_1$                                                                          & H $\times$ $N_2$                                                                       \\ \midrule
Bootstrapping (BCa)     & H $\times$ $N_1$                                                                          & H $\times$ B $\times$ $ N_2$                                                                     \\ \midrule
M-Fold Cross-Validation & \begin{tabular}[c]{@{}l@{}}(H $\times$ M $\times$ $N \times$ (M-1))/M \\ = H $\times$ $N$ $\times$ (M-1)\end{tabular} & \begin{tabular}[c]{@{}l@{}}(H $\times$ M $\times$ $N$ $\times$ 1)/M \\ = H $\times$ N\end{tabular}            \\ \midrule
K-Repeat RRS            & H $\times$ K $\times$ $N_1$                                                                      & H $\times$ K $\times$ $N_2$                                                                   \\ \midrule
\begin{tabular}[c]{@{}l@{}}BVFT \\ \citep{xie2021batch}\end{tabular}                    & H $\times$ $N_1$                                                                          & \begin{tabular}[c]{@{}l@{}}(H $\times$ H) $\times$ $N_2$ or \\ (H $\times$ H) $\times$ $n_2$\end{tabular}         \\ \midrule
\begin{tabular}[c]{@{}l@{}}BVFT-auto \\ \citep{zhang2021towards}\end{tabular}          & H $\times$ $N_1$                                                                          & \begin{tabular}[c]{@{}l@{}}P $\times$ (H $\times$ H) $\times$ $N_2$ or \\ P $\times$ (H $\times$ H) $\times$ $n_2$\end{tabular} \\ \midrule
\begin{tabular}[c]{@{}l@{}}BVFT-FQE \\ \citep{zhang2021towards}\end{tabular}          & H $\times$ $N_1$                                                                          & \begin{tabular}[c]{@{}l@{}} P $\times$ H $\times$ (F $\times$ F) $\times$ $N_2$ or \\ P $\times$ H $\times$ (F $\times$ F) $\times$ $n_2$\end{tabular} \\ 
\bottomrule
\end{tabular}
\end{table}
For BVFT, one can amortize the computational cost by caching (storing $Q(s, a)$ for all $(s, a)$ in the dataset). If caching is done only once, we treat the actual computation time for the validation data set as $n_2$. P is usually between 5 and 10.
When H is relatively large, for example, H = 540 (in our experiment), H * H = 2.916e5. It’s easy to see that RRS is slightly more expensive than M-Fold CV but less expensive than the pairwise comparison tournament algorithm (BVFT). 
\cite{zhang2021towards} proposed BVFT-FQE that only makes pairwise tournament comparison between FQE hyperparameters -- F is 5 in our experiments.
It’s also worth noting that BCa has a high evaluation cost when B is large – when B = 100, BCa evaluation cost is significantly higher than CV and RRS.

\subsection{Sensitivity to OPE Methods}
\label{ap:ope-sensitivity}

OPE is often a critical part of OPL, which has motivated significant research into OPE. Thus the employed OPE method will likely impact the performance of our proposed pipeline. As has been demonstrated in a recent bake-off paper~\citep{voloshin2021empirical},  minimal-assumption OPE methods like weighted doubly robust methods (e.g. \cite{jiang2015dependence}; \cite{thomas2016data}) may be most consistently accurate for many domains. However if the domain is known to be Markov and the models are well specified, FQE methods will likely be more accurate in small data regimes.

To explore further the impact of the choice of OPE method, we conducted an additional experiment on the Sepsis-POMDP domain. 
The aim to was to look at the sensitivity of  \ssrrrs for picking the best \ah to the choice of OPE estimators. In addition to the prior OPE methods used in the main text, we included clipped IS (importance sampling), CWPDIS~\citep{thomas2016data}, and 8 different FQE OPE variants, in which different networks, learning rate and epochs were used.

\begin{table}[h]
\centering
\begin{tabular}{@{}rcc@{}}
\toprule
Sepsis-POMDP   & Parameters                   & \begin{tabular}[c]{@{}c@{}}Best \ah Performance\\ Chosen by\\ \ssrrrs K=5\end{tabular} \\ \midrule
FQE-1          & {[}64{]}, lr=3e-4, epoch=20  & 2.84                                                                                \\
FQE-2          & {[}64{]}, lr=1e-5, epoch=20  & -74.26                                                                              \\
FQE-3          & {[}64{]}, lr=3e-4, epoch=50  & -20.88                                                                              \\
FQE-4          & {[}64{]}, lr=1e-5, epoch=50  & -14.16                                                                              \\
FQE-5          & {[}128{]}, lr=3e-4, epoch=20 & -75.26                                                                              \\
FQE-6          & {[}128{]}, lr=1e-5, epoch=20 & -14.48                                                                              \\
FQE-7          & {[}128{]}, lr=3e-4, epoch=50 & -75.54                                                                              \\
FQE-8          & {[}128{]}, lr=1e-5, epoch=50 & -74.26                                                                              \\ \midrule
IS             & N/A                          & 4.47                                                                                \\
CWPDIS & N/A                          & 4.68                                                                                \\
WIS            & N/A                          & 6.75                                                                                \\ \bottomrule
\end{tabular}
\vspace{2mm}
\caption{Using different OPE estimators in the \ssrrrs pipeline. FQE-1 denotes the FQE with the optimal FQE hyperparameter (heuristically chosen).}
\end{table}

 First, using FQE does generally much worse in this setting which is not very surprizing: FQE assumes the domain is Markov, which Sepsis-POMDP is not.
 
 All importance-sampling based OPE methods yield quite similar performing algorithm-hyperparameter choices in this setting. 
 
 While there are some clear differences, if some basic information about the domain is known (Markov or not), it is likely possible to select a pretty good OPE. In addition, prior work has proposed heuristics~\citep{voloshin2021empirical} or automatic methods for  automatic OPE selection~\citep{su2020adaptive,lee2021model}. An interesting direction for future work would be to include such methods in the pipeline. 


We highlight that while it is well known that OPE methods are important, our paper focused on an under-explored issue: that the dataset partitioning can also introduce a substantial amount of \textit{additional} impact on learning good policies / selecting good \ah.

\subsection{Robustness of \ssrrrs}
\label{ap:robustness}

In Table~\ref{tab:alldomains}, we only show the performance of the best policy among all \ah pairs.
Here we show that \ssrrrs can still robustly select a good hyperparameter for a given offline RL policy learning algorithm (the gap between best \ah selected and true best \ah is relatively small).
\begin{table}[h]
\begin{tabular}{@{}rcccc@{}}
\toprule
Sepsis-POMDP & \begin{tabular}[c]{@{}c@{}}Range of True \\ Policy Performance\\ (95\%CI)\end{tabular} & \begin{tabular}[c]{@{}c@{}}Percentile of \ah \\ Chosen by \ssrrrs \end{tabular} & \begin{tabular}[c]{@{}c@{}}Performance of \ah \\ Chosen by \ssrrrs\end{tabular} & \begin{tabular}[c]{@{}c@{}}True Best \ah \\ Performance\end{tabular} \\ \midrule 
BCQ          & {[}-10.8, -0.73{]}                                                                     & 94\%                                                                       & 5.98                                                                             & 7.86                                                                \\
MBSQI        & {[}-7.34, -2.26{]}                                                                     & 95\%                                                                       & 6.40                                                                             & 7.42                                                                \\
BC           & {[}-8.98, -8.37{]}                                                                     & 58\%                                                                       & -8.46                                                                            & -7.42                                                               \\
BC+PG      & {[}-5.55, -4.26{]}                                                                     & 78\%                                                                       & -3.68                                                                            & 2.52                                                                \\
P-MDP        & {[}-31.17, -21.26{]}                                                                   & 83\%                                                                       & 0.23                                                                             & 2.82                                                                \\ \bottomrule
\end{tabular}
\vspace{2mm}
\caption{We show the relative position (percentile) of the \ah selected by \ssrrrs K=5 pipeline.}
\end{table}

For each algorithm, we evaluate over 24 to 72 hyperparameters, and we compute the 95\% confidence interval of all these policies’ true performance. 
Except for behavior cloning, we are picking hyperparameters that are out-performing 78\%-95\% of other hyperparameters in the same algorithm.

\subsection{Is FQE biased towards FQI algorithms?}
\label{ap:fqe-fqi-bias}

In our evaluation on the Sepsis domain, FQE is used to evaluate both BCQ and MBSQI (both FQI-based) and BC and BCPG (policy-gradient algorithms).

We designed the following analysis experiment using our logged results. We first rank all AH pairs (540 of them) with their true performance in the simulator, and then we count the percentage of FQI (BCQ, MBSQI) algorithms that appear in the top 10\%, 20\%, and 50\% percentile. The number in each cell should be read as: ``90.7\% of AH pairs in the top-10\% based on True Performance are FQI-based''. If FQE is biased towards FQI algorithms, we expect to see a higher percentage of BCQ and MBSQI AH pairs selected than the true performance baseline and compared to other OPE methods.

\begin{table}[h]
\centering
\begin{tabular}{@{}rcc@{}}
\toprule
\begin{tabular}[c]{@{}r@{}}Sepsis-POMDP\\ OPE Method\end{tabular} & \begin{tabular}[c]{@{}c@{}}\% of BCQ and MBSQI \\ AHs in Top-10\% AHs\end{tabular} & \begin{tabular}[c]{@{}c@{}}\% of BCQ and MBSQI \\ AHs in Top-20\% AHs\end{tabular} \\ \midrule
True Performance                                                  & 90.7\%                                                                             & 61.1\%                                                                             \\ \midrule 
FQE-1                                                             & 0\%                                                                                & 0\%                                                                                \\
WIS                                                               & 9.4\%                                                                              & 35.5\%                                                                             \\
RRS-5 WIS                                                         & 68.5\%                                                                             & 58.3\%                                                                             \\ \bottomrule
\end{tabular}
\vspace{2mm}
\caption{Examining whether FQE as an estimator will prefer FQI policy learning algorithms.}
\end{table}
Based on this analysis, we believe that FQE is not biased to select FQI-based algorithms in the Sepsis-POMDP domain. However, our analysis is limited to one domain and only on two FQI-based algorithms. Further investigation is needed but beyond the scope of our paper.

\subsection{Additional Discussions}
\label{ap:add-discussion}

\paragraph{Sensitivity to K in small and large datasets} 
In general, we expect the issue of data partitioning into a train and test split is most important in small datasets: as the dataset gets very large, a single train/test split will generally work well. Therefore, we suggest using a larger K for smaller datasets, but for larger datasets, a smaller K will likely be sufficient. Using our theoretical example in the appendix (chain-MDP), this can also be observed – with a larger N, the failure probability for smaller numbers of repeats decreases. This N-K tradeoff has computational benefits if there is a limited computational budget (larger datasets will require more training, therefore, harder to use a larger K).

\paragraph{Weighted importance sampling (WIS) as a biased estimator} WIS is a self-normalizing importance sampling estimator. We refer readers to \cite{owen2013monte} Chapter 9 for a more detailed discussion on the statistical properties of this type of estimator. In Section~\ref{sec:theory} (line 174), we state:

\begin{displayquote}
WIS will return the observed return of the behavior policy if averaging over a single trajectory, independent of the target policy to be evaluated.
\end{displayquote}

In brief, WIS works by first computing the probability of the dataset trajectory appearing under the evaluation policy and behavior policy:
$$w_i = \prod^L_{t=1} \frac{\pi_e(a_t \vert s_t)}{\pi_b(a_t \vert s_t)}$$
Then, this coefficient is normalized before multiplying with the trajectory return, therefore:
$$\text{WIS}(D) = \frac{1}{n} \sum_{i=1}^n \frac{w_i}{\sum_{j=1}^n w_j} (\sum_{t=1}^L \gamma^t R_t^i).$$
Perhaps surprisingly, if there is a single trajectory, $n=1$, this implies 
$$\text{WIS}(D) =\frac{w_i}{w_i} (\sum_{t=1}^L \gamma^t R_t^i) = \sum_{t=1}^L \gamma^t R_t^i.$$
Here WIS is a biased estimator that returns the trajectory weighted reward, independent of $w_i$.

\subsection{Additional Experiment}

We report the D4RL HalfCheetah result over the same setting as D4RL Hopper, where the result is averaged over 20 runs.

\begin{table*}[ht!]
\centering
\small
\begin{tabular}{rccccccc}
\toprule
\begin{tabular}[c]{@{}r@{}}Re-trained\\ on full dataset\end{tabular} & \begin{tabular}[c]{@{}c@{}}BVFT\\ $\pi$ x FQE\end{tabular} & \begin{tabular}[c]{@{}c@{}}BVFT\\ $\pi$ + FQE\end{tabular} & \begin{tabular}[c]{@{}c@{}} CV-2\end{tabular} & \begin{tabular}[c]{@{}c@{}} CV-5\end{tabular} & \begin{tabular}[c]{@{}c@{}}\ssr\\ RRS-2\end{tabular} & \begin{tabular}[c]{@{}c@{}}\ssr\\ RRS-5\end{tabular} & \multicolumn{1}{c}{\begin{tabular}[c]{@{}c@{}}Optimal\\ Policy\end{tabular}} \\ \midrule
\multicolumn{1}{l}{\textbf{D4RL (HalfCheetah):}} &  &  &  &  &  &  &  \\
random & -1.14  & 1106.94  & -1.13 & -1.13 & \textbf{1922.07} & \textbf{1922.07} & 1922.07 \\
medium & \textbf{4421.95}  & 4290.33 & 4290.33 & 4290.33 & 4290.33 & 4290.33 & 4517.96 \\
medium-expert & 8118.84 & 8799.66  & 8118.84 & 8118.84 & \textbf{9681.78} & \textbf{9681.78} & 10364.36 \\ \bottomrule
\end{tabular}
\caption{Additional comparison of the performance obtained by a policy deployed using the \ssr pipeline vs. using 1-split policy selection approaches on D4RL HalfCheetah. Cells = average true return.}
\label{tab:half-cheetah}
\end{table*}

\subsection{Figure Generation Procedure}
\label{ap:figgen}
Given our partition generation procedure, there are some methods (One-Split OPE, $K \times 2$ Nested CV, and \ssrrrs $K$ when $K < 5$) that have a few different partitions to choose from. For example, out of the 5 RRS split partitions, which partition should we choose for the One-Split OPE method? If we choose one partition, and the One-Split method cannot select the best algorithm, does that mean the One-Split method is bad, or could the 9 other partitions do better for the One-Split method? In order to evaluate these approaches fairly, we  exhaustively train and evaluate on the 5 RRS splits, swap the train/valid set, and train/evaluate on them again, generating 20 scores. For the aforementioned methods, we randomly sample from these 10 (or 20, if Nested CV is being evaluated) scores to simulate the setting that we happen to get one particular split. We run this sampling procedure multiple times and compute the average performance of the policies that are chosen by conditioning on one or $K$ particular partitions. 



\subsection{Domain Descriptions}
\label{ap:domains}
\paragraph{Sepsis.}
The first domain is based on the simulator and works by \cite{oberst2019counterfactual} and revolves around treating sepsis patients. The goal of the policy for this simulator is to discharge patients from the hospital. There are three treatments the policy can choose from antibiotics, vasopressors, and mechanical ventilation. The policy can choose multiple treatments at the same time or no treatment at all, creating 8 different unique actions. 

The simulator models patients as a combination of four vital signs: heart rate, blood pressure, oxygen concentration and glucose levels, all with discrete states (for example, for heart rate low, normal and high). There is a latent variable called diabetes that is present with a $20\%$ probability which drives the likelihood of fluctuating glucose levels. When a patient has at least 3 of the vital signs simultaneously out of the normal range, the patient dies. If all vital signs are within normal ranges and the treatments are all stopped, the patient is discharged. The reward function is $+1$ if a patient is discharged, $-1$ if a patient dies, and $0$ otherwise.

We follow the process described by \cite{oberst2019counterfactual} to marginalize an optimal policy's action over 2 states: glucose level and whether the patient has diabetes. This creates the \textbf{Sepsis-POMDP} environment. We sample 200, 1000, and 5000 patients (trajectories) from Sepsis-POMDP environment with the optimal policy that has 5\% chance of taking a random action. We also sample 200 trajectories from the original MDP using the same policy; we call this the \textbf{Sepsis-MDP} environment.

\paragraph{Robomimic.}
Our approach is further evaluated on a third domain, Robomimic~\citep{mandlekar2021what}, consisting of various continuous control robotics environments along with corresponding sets of suboptimal human data. More specifically, we use the \textbf{Can-Paired} dataset composed of mixed-quality human data. These 200 demonstrations include an equal combination of “good” (the can is picked up and placed in the correct bin) and “bad” trajectories (the can is picked up and thrown out of the robot workspace). The initializations of the tasks being identical, it is expected that algorithms dealing with suboptimal data will be able to filter out the good trajectories from the bad ones and achieve near-optimal performance. Interestingly, state-of-the-art batch RL algorithms do not reach maximum performance~\citep{mandlekar2021what}, making this task a good testbed for our procedure. We also use the \textbf{Transport} dataset, where two robot arms must transfer an object from one bin to another. The dataset contains 200 successful trajectories collected by one human operator.

\paragraph{D4RL.}
\label{ap:d4rl}
D4RL~\citep{fu2020d4rl} is an offline RL standardized benchmark designed and commonly used to evaluate the progress of offline RL algorithms. We use 3 datasets of different quality from the Hopper task: \underline{hopper-random} with 200k samples from a randomly initialized policy, \underline{hopper-medium} with 200k samples from a policy trained to approximately 1/3 the performance of a policy trained to completion with SAC ("expert"), and \underline{hopper-medium-expert} with 200k samples from a 50-50 split of medium and expert data. The Hopper task is to make a hopper with three joints, and four body parts hop forward as fast as possible.

\subsection{TutorBot Domain}
\label{ap:tutorbot}

We introduce a new TutorBot simulator that is designed to mimic 3-5th grade elementary school children in understanding the concept of calculating volume, and engaging them while doing so. We base certain aspects of this simulator on some experimental studies of this learning environment, where an RL policy can learn to teach.
The state space includes children's pre-test score, anxiety level, thinking time, and whether it's the last question in the tutoring session. The action is to offer encouragement, a guided prompt, or a hint at each step of the tutoring. 

\begin{table}[h]
\centering
\begin{tabular}{@{}ccc@{}}
\toprule
  TutorBot  & \begin{tabular}[c]{@{}c@{}}Dimension \end{tabular}  & \begin{tabular}[c]{@{}c@{}}Description \end{tabular}\\ \midrule
\begin{tabular}[c]{@{}c@{}}State \end{tabular} & 4     & \begin{tabular}[c]{@{}c@{}} Pre-test $\in \{0, 1, ..., 8\}$, Anxiety-level $\in [-1, 0]$ \\ Thinking $\in [0, 1]+$, Pre-termination $\in \{0, 1\}$  \end{tabular}                                             \\ \midrule
\begin{tabular}[c]{@{}c@{}}Action\end{tabular} &  1   &         \begin{tabular}[c]{@{}c@{}} 0 = Encourage, 1 = Guided Prompt, 2 = Hint \end{tabular}                                       \\ \midrule 
\begin{tabular}[c]{@{}c@{}}Reward\end{tabular} &  1 & \begin{tabular}[c]{@{}c@{}}0 for all steps if not last step \\ \end{tabular}\\
\bottomrule
\end{tabular}
\vspace{2mm}
\caption{MDP specification for TutorBot.}
\end{table}

The dynamics of TutorBot is a 4th-order Markov transition function that takes in anxiety and the amount of thinking time and updates a latent parameter that captures learning progress. For each simulated student learning trajectory, we pre-determine how many times this student will interact with the TutorBot. We denote this as $T$, which is the trajectory length. We calculated the relationship between $T$ and the pre-test score based on the aforementioned experimental study.
$$T = \text{round}(7 - 0.46 * \text{pre-test} + l), l \sim U([-1, 2])$$
$$\theta_x = [0, -0.05, -0.2, -0.5], \theta_h = [0.5, 0.3, 0.2, 0]$$
$$T(s_{t+1}|s_t, a_t) = \Big[\text{pre-test}, [s_{t-3}, s_{t-2}, s_{t-1}, s_t] \theta_x^T, [s_{t-3}, s_{t-2}, s_{t-1}, s_t] \theta_h^T, \mathbbm{1}\{t+1 = T\} \Big]$$
The reward is always 0 at all steps except for the final step. We use $x$ to denote anxiety and $h$ to denote thinking. Note that anxiety is always negative. We calculate the final reward as follows:
$$R_T = \mathbbm{1}\{U[0, 1] < p\} * r_{\text{improv}} + (1 - \mathbbm{1}\{U[0, 1] < p\}) * r_{\text{base}}, p = x + h$$
Under this simulator, a student will improve a small amount even if the chatbot fails to teach optimally. 
$$r_{\text{improv}} \sim \mathcal{N}(\mu_{\text{improv}}, 1), r_{\text{base}} \sim \mathcal{N}(\mu_{\text{base}}, 0.4)$$
We provide the full simulator code in the GitHub repo.

\subsection{Sepsis-POMDP and Sepsis-MDP Experiment Details}
\label{ap:exp2}

Our algorithm-hyperparameter search is trying to be as realistic as possible to the setting of offline RL practitioners. We search over hyperparameters that could potentially have a strong influence on the downstream performance. Since this is an offline RL setting, we are particularly interested in searching over hyperparameters that have an influence on how pessimistic/conservative the algorithm should be. 

\subsubsection{BCQ}
Batch Constrained Q-Learning (BCQ) is a commonly used algorithm for batch (offline) reinforcement learning~\citep{fujimoto2019off}. We search over the following hyperparameters:

\begin{table*}[h]
\centering
\begin{tabular}{@{}cc@{}}
\toprule
  BCQ   & \begin{tabular}[c]{@{}c@{}}Hyperparameter\\ Range\end{tabular} \\ \midrule
\begin{tabular}[c]{@{}c@{}}Actor/Critic \\ network \\ dimension\end{tabular} & [32, 64, 128]                                                  \\ \midrule
\begin{tabular}[c]{@{}c@{}}Training \\ Epochs\end{tabular}                   & [15, 20, 25]                                                   \\ \midrule
\begin{tabular}[c]{@{}c@{}}BCQ\\ Threshold $\delta$\end{tabular}                      & [0.1, 0.3, 0.5]                                                \\ \bottomrule
\end{tabular}
\caption{BCQ Hyperparams for Spesis-POMDP N=200, 1000. Sepsis-MDP N=200. TutorBot N=200.}
\end{table*}

BCQ threshold determines if the Q-network can take the max over action to update its value using $(s, a)$ -- it can only update Q-function using $(s, a)$ if $\mu(s) > \delta$ and $\pi(a|s) > 0$. The higher $\delta$ (BCQ threshold) is, the less data BCQ can learn from. $\delta$ determines whether $(s', a') \in \mathcal{B}$.
\begin{equation}
\begin{split}
Q(s, a) \leftarrow &(1 - \alpha)Q(s, a) \\ &+ \alpha(r+ \gamma \max_{a' \text{s.t.} (s', a') \in \mathcal{B} }Q'(s', a'))  
\end{split}
\end{equation}

We search through the cross-product of these, in total 27 combinations.

For Sepsis-POMDP N=5000, we realize the network size is too small to fit a relatively large dataset of 5000 patients. So we additionally search over Table~\ref{tab:bcq_n5000}. The actor/critic network uses a 2-layer fully connected network. This resulted in 6 additional combinations for BCQ in Sepsis-POMDP N=5000.


\begin{table*}[h]
\centering
\begin{tabular}{@{}cc@{}}
\toprule
  BCQ   & \begin{tabular}[c]{@{}c@{}}Hyperparameter\\ Range\end{tabular} \\ \midrule
\begin{tabular}[c]{@{}c@{}}Actor/Critic \\ network \\ dimension\end{tabular} & [256, 256],  [512,512], [1024,1024]                                               \\ \midrule
\begin{tabular}[c]{@{}c@{}}Training \\ Epochs\end{tabular}                   &  [25]                                                   \\ \midrule
\begin{tabular}[c]{@{}c@{}}VAE Latent \\ Dim\end{tabular} &  [512] \\ \midrule
\begin{tabular}[c]{@{}c@{}}BCQ\\ Threshold $\delta$\end{tabular}                      & [0.3, 0.4]                                                \\ \bottomrule
\end{tabular}
\caption{BCQ Hyperparams for Spesis-POMDP N=5000.}
\label{tab:bcq_n5000}
\end{table*}



\subsubsection{MBS-QI}
The MBS-QI algorithm is very similar to BCQ, but MBS-QI also clips the states~\citep{liu2020provably}.
We searched through similar hyperparameters as BCQ. 

\begin{table*}[h]
\centering
\begin{tabular}{@{}cc@{}}
\toprule
  MBS-QI   & \begin{tabular}[c]{@{}c@{}}Hyperparameter\\ Range\end{tabular} \\ \midrule
\begin{tabular}[c]{@{}c@{}}Actor/Critic \\ network \\ dimension\end{tabular} & [32, 64, 128]                                                  \\ \midrule
\begin{tabular}[c]{@{}c@{}}Training \\ Epochs\end{tabular}                   & [15, 20, 25]                                                   \\ \midrule
\begin{tabular}[c]{@{}c@{}}BCQ\\ Threshold $\delta$\end{tabular}                      & [0.1, 0.3, 0.5]                                                \\ \midrule 
\begin{tabular}[c]{@{}c@{}}Beta $\beta$ \end{tabular}                      & [1.0, 2.0, 4.0]                                                \\
\bottomrule
\end{tabular}
\caption{MBS-QI Hyperparams for Spesis-POMDP N=200, 1000. Sepsis-MDP N=200. TutorBot N=200.}
\end{table*}

The beta ($\beta$) hyperparameter in MBS-QI is a threshold for the VAE model's reconstruction loss. When the reconstruction loss of the next state is larger than beta, MBS-QI will not apply the Q function on this next state to compute future reward (to avoid function approximation over unfamiliar state space).

\begin{equation}
\begin{split}
&\zeta(s, a ; \widehat{\mu}, b) = \mathbbm{1}(\widehat{\mu}(s, a) \geq \beta) \\
&(\tilde{\mathcal{T}} f)(s, a) := r(s, a) + \gamma \E_{s'} [\max_{a'} \zeta \circ f(s', a')] 
\end{split}
\end{equation}

We search through the cross-product of these, in total 81 combinations. Similar to BCQ situation, we realize the network size is too small to fit a relatively large dataset of Sepsis-POMDP N=5000. So we additionally search over Table~\ref{tab:mbsqi_n5000}. The actor/critic network uses a 2-layer fully connected network. This results in 18 additional combinations for MBS-QI in Sepsis-POMDP N=5000.


\begin{table*}[h]
\centering
\begin{tabular}{@{}cc@{}}
\toprule
  MBS-QI   & \begin{tabular}[c]{@{}c@{}}Hyperparameter\\ Range\end{tabular} \\ \midrule
\begin{tabular}[c]{@{}c@{}}Actor/Critic \\ network \\ dimension\end{tabular} & [256, 256],  [512,512], [1024,1024]                                               \\ \midrule
\begin{tabular}[c]{@{}c@{}}Training \\ Epochs\end{tabular}                   &  [25]                                                   \\ \midrule
\begin{tabular}[c]{@{}c@{}}VAE Latent \\ Dim\end{tabular} &  [512] \\ \midrule
\begin{tabular}[c]{@{}c@{}}BCQ\\ Threshold $\delta$\end{tabular}                      & [0.3, 0.4]                                                \\ \midrule
\begin{tabular}[c]{@{}c@{}}Beta $\zeta$ \end{tabular}                      & [1.0, 2.0, 4.0]                                                \\
\bottomrule
\end{tabular}
\caption{MBS-QI Hyperparams for Spesis-POMDP N=5000.}
\label{tab:mbsqi_n5000}
\end{table*}

\subsubsection{MOPO}

We also experiment with Model-based Offline Policy Optimization (MOPO)~\citep{yu2020mopo}. The original MOPO paper only experimented on Mujoco-based locomotion continuous control tasks. We want to experiment with whether MOPO can work well in environments like the Sepsis-POMDP simulator, which is not only a healthcare domain but also partially observable with a discrete state and action space. We do not expect MOPO to do well. We re-implemented two versions of MOPO with Tensorflow 2.0 and PyTorch, and used the PyTorch version to run our experiments. Our implementation of MOPO matches the original's performance in a toy environment.

MOPO is fairly slow to run -- because it needs first to train a model to approximate the original environment, and then sample from this model to train an RL algorithm. We did not evaluate it for Sepsis-POMDP N=5000.


\begin{table*}[h]
\centering
\begin{tabular}{@{}cc@{}}
\toprule
  MOPO   & \begin{tabular}[c]{@{}c@{}}Hyperparameter\\ Range\end{tabular} \\ \midrule
\begin{tabular}[c]{@{}c@{}} Actor/Critic \\ network \\ dimension \\ dim\end{tabular} & [32, 32], [64, 64], [128, 128]                                                  \\ \midrule
\begin{tabular}[c]{@{}c@{}}Training \\ Iterations \end{tabular}                   & [1000, 2000, 3000]                                                   \\ \midrule
\begin{tabular}[c]{@{}c@{}}MOPO\\ Lambda $\lambda$ \end{tabular}                      & [0, 0.1, 0.2]                                                \\ \midrule 
\begin{tabular}[c]{@{}c@{}}Number of \\ Ensembles \end{tabular}                      & [3, 4, 5]                                                \\
\bottomrule
\end{tabular}
\caption{MOPO hyperparameters for Spesis-POMDP N=200, 1000.}
\end{table*}

Number of ensembles refers to MOPO Algorithm 2, which trains an ensemble of $N$ probabilistic dynamics on batch data. $N$ should be adjusted according to the dataset size. Each dynamics model is trained on $\frac{1}{N}$ of the data during each epoch. 
\begin{equation}
    \widehat T_i(s', r|s, a) = \gN(\mu_i(s, a), \Sigma_i(s, a))
\end{equation}

MOPO $\lambda$ hyperparameter controls how small we want the reward to be, adjusting for state-action pair uncertainty. Generally, the more uncertain we are about $(s, a)$, the more we should ignore the reward that's outputted by the learned MDP model. Its use is also described in Algorithm 2:
\begin{equation}
    \tilde{r}(s, a) \coloneqq r(s, a) - \lambda \max_{i=1}^{N} ||\Sigma_i (s, a) ||_F
\end{equation}

We search through the cross-product of these, in total 81 combinations.

In our initial experiments, MOPO does not seem to perform well in a tabular setting where both state and action are discrete. Therefore, we simplified the idea of MOPO to introduce Pessimistic Ensemble MDP (P-MDP).

\subsubsection{P-MDP}

As noted in the previous section, inspired by MOPO and MoREL~\citep{kidambi2020morel}, we develop a tabular version of MOPO. We instantiate $N$ tabular MDP models. For each epoch, each MDP model only updates on $1/N$ portion of the data. During policy learning time, for each timestep, we randomly sample 1 of the $N$ MDP for the next state and reward; and use Hoeffding bound to compute a pessimistic reward, similar to MOPO's variance penalty on reward:

Let $N(s, a)$ be the number of times $(s, a)$ is observed in the dataset:
\begin{equation}
\begin{split}
     \epsilon &= \beta * \sqrt{\frac{2 \log(1/ \delta)}{N(s, a)}} \\
     \tilde r(s, a) &= \min(\max(r - \epsilon, -1), 1)
\end{split}
\end{equation}

In the last step we bound the reward to (-1, 1) for the Sepsis setting -- but it can be changed to apply to any kind of reward range. We note that Hoeffding bound is often loose when $N(s,a)$ is small, therefore, might make the reward too small to learn any good policy. However, empirically, we observe that in the Sepsis-POMDP, P-MDP is often the best-performing algorithm. We additional add a temperature hyperparameter $\alpha$, that changes the peakness/flatness of the softmax distribution of the learned policy:

\begin{table*}[h]
\centering
\begin{tabular}{@{}cc@{}}
\toprule
  P-MDP   & \begin{tabular}[c]{@{}c@{}}Hyperparameter\\ Range\end{tabular} \\ \midrule
\begin{tabular}[c]{@{}c@{}}Training \\ Iterations \end{tabular}                   & [1000, 5000, 10000]                                                   \\ \midrule
\begin{tabular}[c]{@{}c@{}}Penalty \\ Coefficient $\beta$ \end{tabular}                      & [0, 0.1, 0.5]                                               \\ \midrule 
\begin{tabular}[c]{@{}c@{}}Number of \\ Ensembles \end{tabular}                      & [3, 5, 7]                                                \\ \midrule
\begin{tabular}[c]{@{}c@{}}Temperature $\alpha$ \end{tabular}                      &    [0.05, 0.1, 0.2]                                             \\
\bottomrule
\end{tabular}
\caption{P-MDP Hyperparams for Spesis-POMDP N=200, 1000.}
\end{table*}

Not surprisingly, since planning algorithms (such as Value Iteration or Policy Iteration) need to enumerate through the entire state space, we find it too slow to train a policy in Sepsis-MDP domain, because Sepsis-POMDP has 144 unique states, yet Sepsis-MDP has 1440 unique states (glucose level has 5 unique states and diabetes status has 2 unique states). TutorBot and Robomimic both have continuous state space, therefore are not suitable for our P-MDP algorithm without binning. 

We search through the cross-product of these, in total 81 combinations.

For Sepsis-POMDP N=5000, we realize we can increase the number of MDPs and increase training iterations to fit a relatively large dataset of 5000 patients. So we additionally search over Table~\ref{tab:pmdp_n5000}. This results in 16 additional combinations for P-MDP in Sepsis-POMDP N=5000.

    


\begin{table*}[h]
\centering
\begin{tabular}{@{}cc@{}}
\toprule
  P-MDP   & \begin{tabular}[c]{@{}c@{}}Hyperparameter\\ Range\end{tabular} \\ \midrule
\begin{tabular}[c]{@{}c@{}}Training \\ Iterations \end{tabular}                   & [20000, 40000]                                                   \\ \midrule
\begin{tabular}[c]{@{}c@{}}Penalty \\ Coefficient $\beta$ \end{tabular}                      & [0.05, 0.1]                                               \\ \midrule 
\begin{tabular}[c]{@{}c@{}}Number of \\ Ensembles \end{tabular}                      & [15, 25]                                                \\ \midrule
\begin{tabular}[c]{@{}c@{}}Temperature $\alpha$ \end{tabular}                      &    [0.01, 0.05]                                             \\
\bottomrule
\end{tabular}
\caption{P-MDP Hyperparams for Spesis-POMDP N=5000.}
\label{tab:pmdp_n5000}
\end{table*}

\subsubsection{BC}
Behavior Cloning (BC) is a type of imitation learning method where the policy is learned from a data set by training a policy to clone the actions in the data set. It can serve as a great initialization strategy for other direct policy search methods which we will discuss shortly. 

One pessimistic hyperparameter we can introduce to behavior cloning is similar in spirit to BCQ and MBS-QI, we can train BC policy only on actions that the behavior policy has a high-enough probability to take, optimizing the following objective:
\begin{equation}
\begin{split}
    \zeta &= \pi_b(a|s) \geq \alpha \\
    \arg \min_\theta &E_{(s,a) \sim D}||\pi_\theta (s) - \zeta \circ \pi_b(a|s) || ^2
\end{split}
\end{equation}
We refer to $\alpha$ as the ``safety-threshold''. We search through the cross-product of these, in total 27 combinations.

\begin{table*}[h]
\centering
\begin{tabular}{@{}cc@{}}
\toprule
  BC   & \begin{tabular}[c]{@{}c@{}}Hyperparameter\\ Range\end{tabular} \\ \midrule
\begin{tabular}[c]{@{}c@{}}Policy network \\ dimension\end{tabular} & [32, 32], [64, 64], [128, 128]                                              \\ \midrule
\begin{tabular}[c]{@{}c@{}}Training \\ Epochs\end{tabular}                   & [15, 20, 25]                                                   \\ \midrule
\begin{tabular}[c]{@{}c@{}}Safety\\ Threshold $\alpha$ \end{tabular}                      & [0, 0.01, 0.05]                                                \\ \bottomrule
\end{tabular}
\caption{BC Hyperparams for Spesis-POMDP N=200, 1000, 5000. Sepsis-MDP N=200. TutorBot N=200.}
\end{table*}



\subsubsection{POIS}
Policy Optimization via Importance Sampling~\citep{metelli2018policy} uses an importance sampling estimator as an end-to-end differentiable objective to directly optimize the parameters of a policy. In our experiment, we refer to this as the ``\textbf{PG}'' (policy gradient) method. Similar to BC method, we can set a safety threshold $\alpha$ that zeros out any behavior probability of an action that's not higher than $\alpha$, and then re-normalizes the probabilities of other actions. \cite{metelli2018policy} also introduces another penalty hyperparameter $\lambda$ to control the effective sample size (ESS) penalty. ESS measures the Renyi-divergence between $\pi_b$ and $\pi_e$. 
Let $\widehat V$ be the differentiable importance sampling estimator -- we write the optimization objective similar to \cite{futoma2020popcorn}, but without the generative model:
\begin{equation}
\begin{split}
    \gJ(\train) &= \widehat V(\pi_\theta; \train) - \frac{\lambda_{\mathrm{ESS}}}{\mathrm{ESS}(\theta)} \\
    \theta &= \arg \max_\theta \gJ(\train)
\end{split}
\end{equation}

We search through the following hyperparameters in Table~\ref{tab:pois}. There are 81 combinations in total.

\begin{table*}[h]
\centering
\begin{tabular}{@{}cc@{}}
\toprule
  BC   & \begin{tabular}[c]{@{}c@{}}Hyperparameter\\ Range\end{tabular} \\ \midrule
\begin{tabular}[c]{@{}c@{}}Policy network \\ dimension\end{tabular} & [32, 32], [64, 64], [128, 128]                                              \\ \midrule
\begin{tabular}[c]{@{}c@{}}Training \\ Epochs\end{tabular}                   & [15, 20, 25]                                                   \\ \midrule
\begin{tabular}[c]{@{}c@{}}Safety\\ Threshold $\alpha$ \end{tabular}                      & [0, 0.01, 0.05]                                                \\ \midrule
\begin{tabular}[c]{@{}c@{}}ESS\\ Penalty $\lambda$ \end{tabular}                      & [0, 0.01, 0.05]                                                \\ 
\bottomrule
\end{tabular}
\caption{POIS, BC+POIS, BC+mini-POIS Hyperparams for Spesis-POMDP N=200, 1000, 5000. Sepsis-MDP N=200. TutorBot N=200.}
\label{tab:pois}
\end{table*}



\subsubsection{BC+POIS}
BC + POIS is a method that first finds a policy using BC as an initialization strategy to make sure that the policy stayed close (at first) to the behavior policy. This is particularly useful for neural network-based policy classes, as a form of pre-training using behavior cloning objective. We use the same set of hyperparameters displayed in Table~\ref{tab:pois}, resulting in 81 combinations in total.

\subsubsection{BC+mini-POIS}
In both \cite{metelli2018policy} and \cite{futoma2020popcorn}, the loss is computed on the whole dataset $\train$, which makes sense -- importance sampling computes the expected reward (which requires averaging over many trajectories to have an estimation with low variance). However, inspired by the success of randomized optimization algorithms such as mini-batch stochastic gradient descent (SGD), we decided to attempt a version of BC + POIS with $\widehat V$ over a small batch of trajectories instead of over the entire dataset. Our batch size is 4 (4 trajectories/patients) for Sepsis-POMDP N=200 and 1000, which is very small. However, this strategy seems to be quite successful, resulting in learning high-performing policies competitive with other more principled methods. This can be seen in Figure~\ref{fig:hyperparam} (``BCMINIPG''). 

We leave the exploration of why this is particularly effective to future work, and hope others who want to try POIS style method to include our variant in their experiment. We use the same set of hyperparameters displayed in Table~\ref{tab:pois}, resulting in 81 combinations in total.

\subsection{TutorBot Experiment Details}
\label{ap:exp3}

The details of this environment is shown in the code file in the supplementary material.
We trained BC+POIS, POIS, and BC+mini-POIS on this domain.

\subsection{Robomimic Experiment Details}
\label{ap:exp4}

We refer the reader to \cite{mandlekar2021what} for a full review of the offline RL algorithms used in our experiment. For Robomimic, we include the range of hyperparameters we have considered below:


\begin{itemize}
    \item BC:
    \begin{itemize}
        \item Actor NN dimension: [300,400], [1024,1024]
        \item Training epochs: 600, 2000
        \item GMM actions: 5, 25
    \end{itemize}
    \item BCRNN:
    \begin{itemize}
        \item RNN dimension: [100], [400]
        \item Training epochs: 600, 2000
        \item GMM actions: 5, 25
    \end{itemize}
    \item BCQ:
    \begin{itemize}
        \item Critic NN size: [300,400], [1024,1024]
        \item Training epochs: 600, 2000
        \item Action samples:  [10,100], [100,1000]
    \end{itemize}
    \item CQL:
    \begin{itemize}
        \item Critic NN size: [300,400], [1024,1024]
        \item Training epochs: 600, 2000
        \item Lagrange threshold: 5, 25
    \end{itemize}
    \item IRIS:
    \begin{itemize}
        \item Critic NN size: [300,400], [1024,1024]
        \item Training epochs: 600, 2000
        \item LR critic: 0.001, 0.0001
    \end{itemize}
\end{itemize}
\subsection{D4RL-Hopper Experiment Details}
\label{ap:exp5}


For the D4RL experiments, we include the range of hyperparameters we have considered below:

\begin{itemize}
    \item BCQ:
    \begin{itemize}
        \item Policy NN size: [512,512], [64,64]
        \item LR policy: 0.001, 0.0001
    \end{itemize}
    \item CQL:
    \begin{itemize}
        \item Policy NN size: [256,256,256], [64,64,64]
        \item LR policy: 0.001, 0.0001
    \end{itemize}
    \item AWAC:
    \begin{itemize}
        \item Policy NN size: [256,256,256,256], [64,64,64,64]
        \item LR policy: 0.001, 0.0001
    \end{itemize}
\end{itemize}

For BVFT Strategy 1 $\pi$ x FQE, we use the optimal FQE hyperparameter on all hyperparameters of BCQ, CQL and AWAC. For BVFT Strategy 2 $\pi$ + FQE, we use 4 FQE hyperparameters but only with 4 hyperparameters of BCQ. For RRS and CV, we use the optimal FQE hyperparameter on 4 hyperparameters of BCQ as well.

\subsection{Computing Resources}
For the overall experimental study in this paper, an internal cluster consisting of 2 nodes with a total of 112 CPUs and 16 GPUs was used.

\subsection{Additional Offline RL Sensitivity Study}
\subsubsection{Sensitivity to data splitting: One-Split OPE}
\begin{figure*}[ht!]%
    \centering
    \includegraphics[width=\linewidth]{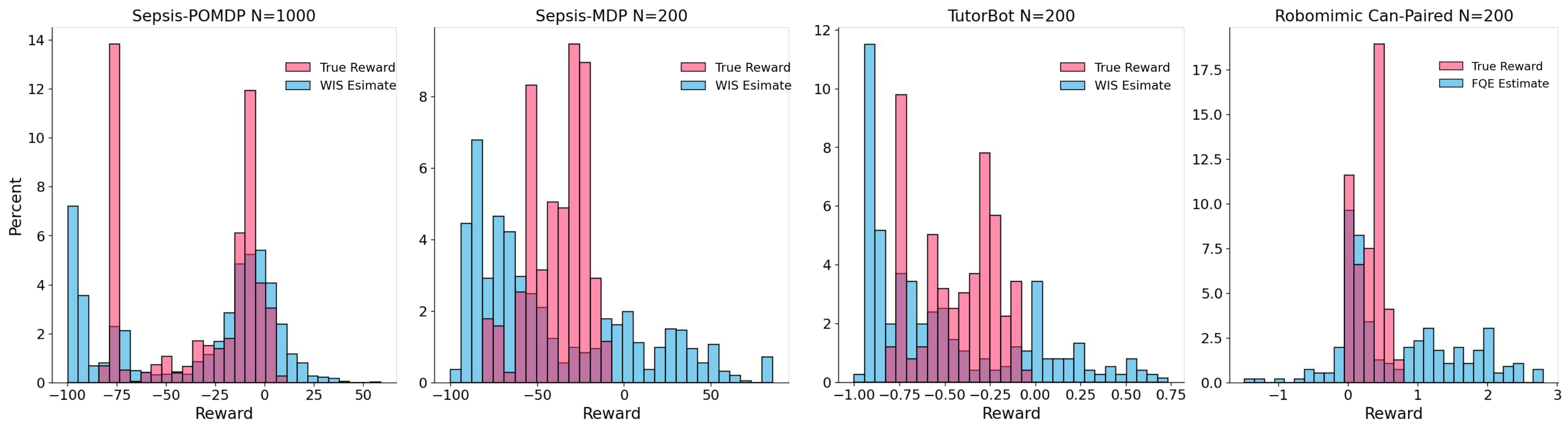}
    \caption{We show that policies learned from offline RL algorithms are sensitive to the variation of training and validation dataset: an algorithm-hyperparameter (AH) pair can obtain wildly different policies based on which portion of the data they were trained on. We obtained 5400 policies from 540 AH combinations on Sepsis-POMDP (N=1000) domain. The variation is not just in terms of the policy's true performance in the real environment, but also in terms of OPE estimations. Note that FQE estimate on Robomimic exceeded the range of possible achievable rewards (between 0 and 1). The true reward is calculated by evaluating the policy in the real environment.}
    \label{fig:variation}
\end{figure*}

Figure~\ref{fig:variation} shows 
that procedure produces policies with drastically different estimated, and true, performances subject to randomness in data selection process. Because training and validation set randomness are directly conflated, it becomes difficult to accurately select a better \ah pair (and its associated higher-performing policy) based on a single train/validation set partition.

\subsubsection{Sensitivity to hyperparameters}
\begin{figure*}[ht!]%
    \centering
    \includegraphics[width=\linewidth]{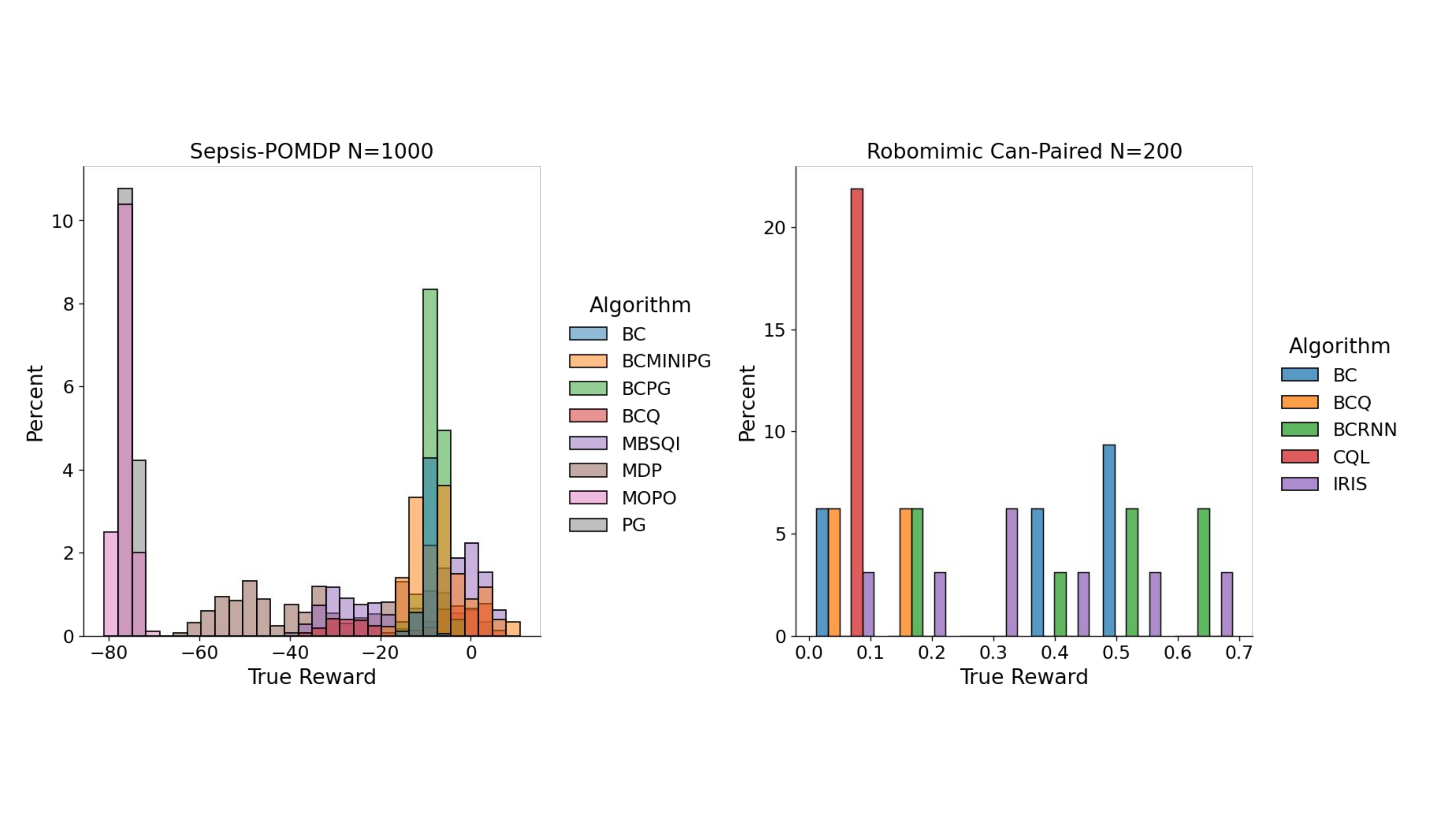}
    \caption{Sensitivity of offline RL algorithms due to the choice of hyperparameters.}
    \label{fig:hyperparam}
\end{figure*}

In Figure~\ref{fig:hyperparam}, we show that offline RL algorithms are sensitive to the choice of hyperparameters. In the Sepsis-POMDP N=1000 task and the Robomimic Can-Paired N=200 task, all popular offline RL algorithms show a wide range of performance differences even when trained on a fixed partition of the dataset.

\end{document}